\documentclass{article} 
\PassOptionsToPackage{hyperfootnotes=false}{hyperref}
\usepackage{arxiv}

\usepackage{amsmath,amsfonts,bm}

\def\eqref#1{equation~\ref{#1}}

\def\ceil#1{\lceil #1 \rceil}

\def\1{\bm{1}}

\DeclareMathAlphabet{\mathsfit}{\encodingdefault}{\sfdefault}{m}{sl}
\SetMathAlphabet{\mathsfit}{bold}{\encodingdefault}{\sfdefault}{bx}{n}

\DeclareMathOperator*{\argmax}{arg\,max}

\usepackage{longtable}

\usepackage{hyperref}
\usepackage{url}
\usepackage[dvipsnames]{xcolor}
\usepackage{subcaption}
\usepackage{cleveref}
\usepackage{enumitem}
\usepackage{amsmath}
\usepackage{graphicx}
\usepackage{makecell}
\usepackage{booktabs}
\usepackage{natbib}
\usepackage{array}
\usepackage[section]{placeins}

\setlist{noitemsep, topsep=2pt, parsep=0pt, partopsep=0pt}
\title{Self-Play Pretraining with Zero Data}

\author{%
  \bfseries
  Aditya Cowsik\,\textsuperscript{1}\thanks{\raggedright Equal contribution; authors listed alphabetically. Correspondence to
    \texttt{aditya.cowsik@gmail.com}, \texttt{kfirdolev@tauex.tau.ac.il},
    \texttt{michaelyli@stanford.edu}, and \texttt{bruno.deluca@lapth.cnrs.fr}.}\quad
  Kfir Dolev\footnotemark[1]\,\textsuperscript{2}\quad
  Michael Y. Li\footnotemark[1]\,\textsuperscript{3}\quad
  G. Bruno De Luca\,\textsuperscript{4}\quad
  Nourya Cohen\,\textsuperscript{2}\\[0.4em]
  \bfseries
  Noah D. Goodman\,\textsuperscript{3}\quad
  Yoav Levine\,\textsuperscript{2}\\[0.7em]
  \normalfont\normalsize
  \textsuperscript{1}Independent Researcher\quad
  \textsuperscript{2}Tel Aviv University\quad
  \textsuperscript{3}Stanford University\quad
  \textsuperscript{4}LAPTh, USMB
  \thanks{\raggedright G. Bruno De Luca, Aditya Cowsik, and Kfir Dolev began this work while affiliated with the Stanford Institute for Theoretical Physics.}
}

\definecolor{RoyalPurple}{RGB}{108, 59, 170}

\begin{document}

\newcommand{\benchmark}{\texttt{mtd\_melody\_16th}}
\newcommand{\code}[1]{\texttt{#1}}
\newcommand{\hashdisplay}[1]{%
  \par\smallskip
  \begingroup
  \centering\ttfamily\scriptsize #1\par
  \endgroup
  \smallskip
}
\newcommand{\NOTE}{\textsc{Note-On}}
\newcommand{\HOLD}{\textsc{Hold}}
\newcommand{\REST}{\textsc{Rest}}
\newcommand{\method}{SP}

\maketitle

\begin{abstract}
Advances in language modeling have been driven by scaling pretraining on ever more data. 
Yet, the training data is still largely curated on the model's behalf.
A more general approach to pretraining would let the model learn to generate the data most useful for its own improvement.
This would provide an effectively unbounded source of training data, limited by compute rather than human knowledge.
We introduce \textbf{Self-Play Pretraining with Zero Data}, an initial proof-of-concept towards realizing this vision. 
Our procedure casts synthetic data generation as a search over the space of all computable structure, taking inspiration from Solomonoff induction.
Starting from random initialization, two models learn in tandem: a generator proposes programs interpreted by a universal Turing machine, generating byte sequences, while a learner autoregressively predicts these byte sequences.
The learner is trained with standard cross-entropy, while the generator is trained with reinforcement learning to produce sequences at the frontier of the learner's capabilities, yielding an adaptive curriculum.
A universal Turing machine gives us a search space over all computable data-generating processes, imposing little domain-specific structure, and  self-play searches over this space for useful training data.
We test whether zero-shot performance on natural data improves predictably with self-play compute; this is a clean test of transfer since neither generator nor learner is trained on natural data.
Across several natural datasets, zero-shot loss exhibits predictable scaling in compute.
The models also exhibit in-context learning, and discover recognizable mathematical sequences during training.
\end{abstract}

\begin{flushright}
\small\itshape
``By teaching, we learn.''
\\\textit{Seneca}
\end{flushright}

\begin{flushright}
\small\itshape
``So much from so little, almost everything from
almost nothing.''
\\\textit{John Archibald Wheeler}
\end{flushright}

\section{Introduction}
\begin{figure}[ht]
    \centering
    \begin{minipage}{\textwidth}
    \centering
        \includegraphics[width=0.75\linewidth]{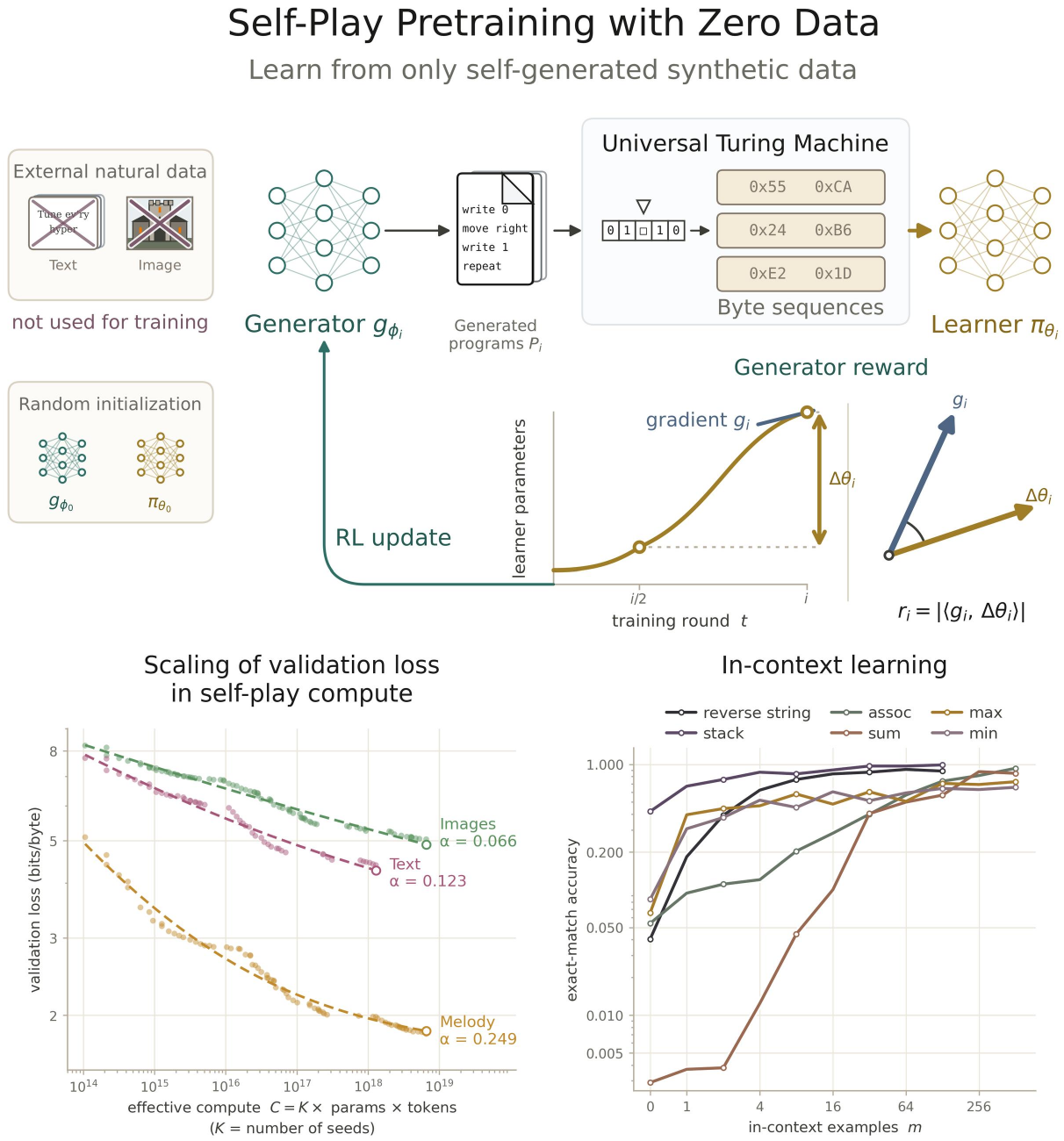}
        \caption[]{\textbf{Self-Play Pretraining with Zero Data}.
Starting from {randomly initialized} models and using only synthetically generated data, self-play produces predictable scaling on held-out natural data. 
\textbf{(Top)}
Our procedure casts synthetic data generation as search over the space of computable structure.
A generator proposes programs, which are executed to produce byte sequences used to train a learner by next-token prediction.
The generator is trained via reinforcement learning to propose programs near the frontier of the learner's capabilities, measured by how strongly the learner's gradients align with the learner's recent learning trajectory with respect to an AdamW-preconditioned inner product.
\textbf{(Lower left)} Self-play produces predictable improvements in validation loss with compute across natural datasets.
We show the compute-optimal frontier over model sizes, number of self-play rounds, and ensemble sizes.
\textbf{(Lower right)} The resulting learner exhibits in-context learning on held-out tasks without any gradient updates. We report empirical success under greedy decoding as a function of the number of in-context examples $m$, averaged over independently sampled task instances.}
        \label{fig:training}
    \end{minipage}
\end{figure}

Advances in language modeling have been driven by pretraining on Internet data.
Yet, the training data is still largely curated and constructed on the model's behalf through large-scale data curation efforts~\citep{li2025datacomplmsearchgenerationtraining, soldaini2024dolmaopencorpustrillion, penedo2023refinedwebdatasetfalconllm}, data mixture design~\citep{xie2023doremioptimizingdatamixtures, chen2026olmixframeworkdatamixing}, and hand-designed synthetic generators~\citep{gunasekar2023textbooksneed,yang2025synthetic}.
A more generic approach to pretraining would let the model itself learn to generate training data that is most useful for its own improvement.
Such an approach would extend the Bitter Lesson~\citep{sutton2019bitter} to the training data itself, minimizing hand-engineered inductive bias and creating a self-contained path to scaling, where compute alone can sustain continued improvement~\citep{kim2026pre, SilverWelcomeTT}.

As an initial step towards this vision, we introduce a self-play algorithm for pretraining from \emph{zero data}. 
Starting from random initialization, two autoregressive transformers co-evolve: a generator proposes programs for a minimal \emph{universal Turing machine} whose execution produces byte sequences, while a learner is trained on those sequences via next-token prediction. 
We use a universal Turing machine to make the space of all synthetic training data as expressive as possible while imposing minimal domain-specific structure: any computable data-generating process can, in principle, be represented as a program.
However, the space of computable structure is enormous, and only a small subset of programs produce sequences that are useful for the learner. 
Moreover, whether a sequence is useful changes as the learner improves. 
This is why a self-play approach that adapts to the learner is natural: rather than specifying useful structure in advance, we let the generator discover which programs are most useful to the learner as training progresses.
To encourage this behavior, the generator is trained with reinforcement learning using a learning-progress reward, shifting probability toward programs whose outputs lie near the frontier of the learner's current capabilities. 
These design choices place our approach in the lineage of classical \emph{universal prediction}~\citep{solomonoff1964formal,graumoya2024learninguniversalpredictors,hutter2000theoryuniversalartificialintelligence,bloem2025universalpretrainingiteratedrandom,merhav98}, which formalizes how induction can be possible when the hypothesis class contains all computable data-generating processes and the learner has unbounded compute: we aim at an efficient computable approximation to universal prediction.

The resulting self-generated data is only valuable insofar as what the learner learns transfers to natural data. 
Our key hypothesis is that self-play over this space of computable data-generating processes can discover generic predictive regularities---such as copying, recursion, and hierarchical composition---that improve prediction on natural data.
Importantly, we hypothesize that these regularities capture structure independent of \emph{contingent information}---the particular facts, symbols, or modality of any one dataset---that can therefore transfer across data-generating processes.
Indeed, prior work on formal, algorithmic, and non-linguistic pretraining distributions provides evidence that such cross-distribution transfer is possible~\citep{graumoya2024learninguniversalpredictors,papadimitriou-jurafsky-2020-learning,hu2025circuitschomskyprepretrainingformal,lee2026traininglanguagemodelsneural}.

We test this hypothesis through a compute-optimal scaling law analysis.
Concretely, we train randomly-initialized transformers at various scales via self-play and evaluate the resulting learners zero-shot on held-out datasets spanning natural language, images, speech,
melodies, DNA, and mathematical sequences.
For each dataset, we construct a compute-optimal frontier over model size, self-play rounds, and ensemble size.
Across these diverse domains, a single family of self-play models exhibits predictable power-law improvements in zero-shot loss with compute, with scaling exponents comparable to those obtained by training directly on natural data.
Importantly, this is a clean test of transfer since we deliberately run this process \emph{tabula rasa}: both models are randomly
initialized and all learner training data is generated through self-play.
Thus, our experiments isolate the effect of our self-play procedure and test whether useful predictive structure can emerge \emph{ex nihilo}.
We also find that the learner develops in-context learning on completely held-out tasks,
and the generator discovers known mathematical sequences.

\section{Self-Play Pretraining with Zero Data}
\label{sec:method}
We introduce a self-play formulation of pretraining that involves searching over the space of computable structure.
Every training sequence is the output of a program executed on a fixed universal Turing machine $U$ 
and these programs are produced by a learned generator.
Our method has two components: (1) a
\textbf{Learner} $\pi_\theta$, an autoregressive language model
trained to predict program outputs, and (2) a \textbf{Generator} $g_\phi$, an
autoregressive language model defined over programs for $U$. Both are transformers with the same architecture, trained from \emph{random initialization}; the generator is trained via reinforcement learning (RL) and the learner is trained via next-token prediction.
Importantly, we consider this \emph{tabula rasa} setup to cleanly test whether self-play can generate structure that transfers to natural data.

Each round of self-play proceeds as follows:
\begin{enumerate}
  \item \textbf{Program generation:} Sample
    $N$ programs from the generator $\{x_i\}_{i=1}^N \sim g_\phi$.
  \item \textbf{Execution:} Run each program on $U$ to obtain output
    sequences $y_i = U(x_i, \omega_i)$, where $\omega_i$ is a random input
    tape.
  \item \textbf{Learner and Generator update:} The learner takes one gradient step on the output sequences, optimizing the standard next-token loss. 
  The generator takes a policy gradient step with a \emph{learning-progress} reward that encourages the generator to propose programs at the frontier of the learner's capabilities.
  In addition, the generator is updated via a supervised fine-tuning objective on existing programs to mitigate catastrophic forgetting and on mutated programs to promote exploration.
  \end{enumerate}

\subsection{Program space}
\label{sec:method-programs}
We would like the generator's search space to be as expressive as possible while
imposing little domain-specific structure.
We therefore use programs for a minimal universal Turing machine as the substrate
for generating synthetic data.
Specifically, we use a Brainf*ck-like Turing-complete language, following
\citet{graumoya2024learninguniversalpredictors}.
Its primitive instructions manipulate a byte-valued tape, implement loops, and
read or emit bytes.
Because the language is universal, any computable data-generating process can,
in principle, be represented as a program.

Let $x \in \mathcal{A}^{\le L}$ denote a program generated over the machine's
instruction alphabet.
Executing $x$ on the universal machine $U$ with random input tape $\omega$
produces a byte sequence
\[
    y = U(x,\omega) \in \{0,\ldots,255\}^{T}.
\]
The random input tape allows a single program to represent a distribution over
output sequences.
These output bytes, rather than the programs themselves, constitute the
learner's training data.
We design the execution semantics so that every generated string is executable:
programs cannot fail through syntax or memory errors, and execution always
produces a bounded-length output.
We defer the precise execution semantics and resource limits to
Appendix~\ref{apx:turing-machine}.

\subsection{Objectives}
\label{sec:method-objectives}

\paragraph{Program pool.}
At each self-play round $e$, we construct a pool of programs
\[
    \mathcal B_e
    =
    \mathcal B_e^{\mathrm{fresh}}
    \mathbin{\dot\cup}
    \mathcal B_e^{\mathrm{mut}}
    \mathbin{\dot\cup}
    \mathcal B_e^{\mathrm{replay}},
\]
containing fresh samples from the current generator, local mutations of
previously high-reward programs, and programs replayed from earlier rounds.
Fresh samples provide global exploration, mutations refine promising regions of
program space, and replay preserves useful structures discovered earlier in
training.
We write $M_e = |\mathcal B_e|$ for the number of programs in the pool.
Details for mutation, replay, and the program bank are offered in
Appendix~\ref{sec:pool_construction}.

\paragraph{Learner Update.}
The learner is trained by standard next-token prediction on program outputs. For an output sequence $y$, define the
per-sequence loss $\mathcal{L}(y;\theta)$ as the mean cross-entropy over its content tokens.
If $y_i = U(x_i,\omega_i)$ is the output obtained by executing program $x_i$,
the learner objective in round $e$ is
\begin{equation}
  \label{eq:learner-objective}
  \mathcal{L}_{\mathrm{learner}}(\theta;\mathcal{B}_e)
  =
  \frac{1}{M_e}
  \sum_{i\in\mathcal{B}_e}
  \mathcal{L}(y_i;\theta).
\end{equation}
Thus fresh, mutated, and replay programs all train the learner. 

\paragraph{Generator Reward.}
The generator's reward must be capable of identifying programs with useful structure from programs without any external feedback. 
Initially, we considered a reward based on how difficult a sequence is to predict, motivated by prior work on self-play~\citep{pmlr-v267-dong25h, bailey2026scaling}.
However, this has a fundamental failure mode: a program can be made arbitrarily difficult without containing useful structure—for example, by injecting random bytes into an otherwise predictable sequence.

To avoid this degeneracy, we evaluate a new program based on whether it builds on what the learner has actually been able to learn. 
Intuitively, the learner's change in parameters summarizes this: learning signals arising from reusable structure accumulate, whereas we expect that idiosyncratic effects that are not learnable do not. 
Concretely, we reward the generator for producing programs whose learner gradients align with the
learner's current learning trajectory. Let
\[
  p(e)=\lfloor e/2\rfloor,
  \qquad
  \delta\theta_e=\theta_{p(e)}-\theta_e,
\]
where $\theta_{p(e)}$ is the learner checkpoint at the lookback horizon. The
reward for program $x_i$ with output $y_i$ is
\begin{equation}
  \label{eq:reward}
  r_i
  =
  |\left\langle
    \nabla_\theta \mathcal{L}(y_i;\theta_e),
    P_e\odot \delta\theta_e
  \right\rangle|,
\end{equation}
where
\[
  P_e
  =
  \frac{\mathrm{lr}}{\sqrt{\hat v_e}+\epsilon}
\]
is the diagonal AdamW step operator obtained from the learner's optimizer
state.
We use the lookback window of $\ceil{e/2}$ so that signals which take a long time to appear can be measured. The growing window  helps average over short-term fluctuations and produce a signal which becomes more stable as training progresses, but allows early mistakes to eventually be forgotten.

We provide some additional intuition below, but we emphasize that we selected this reward after searching through several possibilities at small scale. 
Detailed analysis can be found in \cref{tab:reward_ablations}. 
Formally, this reward is a preconditioned gradient-alignment score, between the learner's gradient on a program's output and the learner's parameter movement over the lookback window, with the diagonal AdamW preconditioner defining the inner product; we found that using this pre-conditioning was important in line with~\citet{thrush2026synthetic}.
Intuitively, this reward favors programs that are not yet mastered, but whose structure extends what the learner has already shown it can learn. 
We expect that programs that are already mastered induce nearly zero gradients, while programs containing unrelated or unlearnable structure induce gradients that do not align with the learner's parameter movement.
Both receive little reward. 
Instead, high reward is assigned to programs that induce substantial learning, but only in directions congruent with the learner's recent progress; this concentrates the generator on the frontier of the learner's current capabilities.
We avoid materializing full gradients by using forward mode automatic differentiation~\citep{griewank2008evaluating} to calculate Equation~\ref{eq:reward}.\footnote{Forward mode kernel implemented in \url{https://github.com/amorehead/jvp_flash_attention}} 

\paragraph{Policy-Gradient RL Objective.}
We train the generator using a KL-regularized expected reward
\begin{equation}
  \label{eq:gen-objective}
  J_{\mathrm{RL}}(\phi)
  =
  \mathbb{E}_{x\sim g_\phi}[r(x)]
  -
  \beta\,\mathrm{KL}(g_\phi\Vert g_0),
\end{equation}
where $\beta$ is the KL regularization coefficient and $g_0$ is the fixed uniform
program prior defined as
\[
  g_0(x)
  =
  |\mathcal{A}|^{-\ell(x)}.
\]
Here $\ell(x)$ is the number of tokens up to and including the terminating
\texttt{F}. 
This is the natural analog of the Solomonoff prior $2^{-|p|}$~\citep{solomonoff1964formal}, which weights programs according to description length. 
The generator is
initialized near $g_0$ and regularized toward it throughout training.

Since the vanilla policy gradient estimator is high variance, we consider a GRPO (batch-level) based estimator~\citep{Guo_2025}. 
Let $\bar r_e$ and $\sigma_{r,e}$ denote the mean and standard deviation,
respectively, of the rewards $\{r_i\}_{i\in\mathcal B_e}$ over the full
round pool. 
We define the advantage of program $i$ as
\[
  A_i
  =
  \frac{r_i-\bar r_e}{\sigma_{r,e}+\epsilon}
  -
  \beta
  \left(
    \log g_\phi(x_i)-\log g_0(x_i)
  \right).
\]

Since our bank consists of off-policy samples, we use a sequence-level importance ratio correction
$\rho_i
  =
  \frac{g_\phi(x_i)}
       {g_{\phi_{\mathrm{old},i}}(x_i)}$ ~\citep{zheng2025groupsequencepolicyoptimization}.
It is equal to one for fresh
on-policy samples; for replay samples, its denominator is the sampling
probability stored when the program originally entered the replay bank.
\newcommand{\stopgrad}[1]{\operatorname{stopgrad}\!\left(#1\right)}
The policy-gradient term is then
\begin{equation}
  \label{eq:pg-surrogate}
  \mathcal{L}_{\mathrm{PG}}(\phi)
  =
  -\frac{1}{\left|\mathcal{B}_e \setminus \mathcal{B}^{\text{mut}}_e\right|}
  \sum_{i\in\mathcal{B}_e \setminus \mathcal{B}^{\text{mut}}}
  \stopgrad{\rho_i}\,
  \log g_\phi(x_i)\,
  \stopgrad{A_i}.
\end{equation}
Mutation rows are excluded because they were not
sampled from a proposal distribution with a well-defined log probability.
We clip the  sequence-level
importance ratio to ensure $\rho_i\in[e^{-20},e^{20}]$.

\paragraph{Expert Iteration.}
\label{sec:method-expert-iteration}
To prevent forgetting, we replay previous programs by distilling high-reward programs
back into the generator using reward-weighted supervised fine-tuning over the
full pool $\mathcal{B}_e$.
This is a technique used to mitigate forgetting in pretraining and RL~\citep{ibrahim2024simplescalablestrategiescontinually,schaul2016prioritizedexperiencereplay}.
We assign each program a normalized sequence-level weight
\[
w_i
=
\frac{[r_i]_+}{\sum_{j\in\mathcal{B}_e}[r_j]_+},
\qquad
[r]_+ \equiv \max(r,0),
\]
and optimize
\[
\mathcal{L}_{\mathrm{EI}}(\phi;\mathcal{B}_e)
=
-\sum_{i\in\mathcal{B}_e} w_i \log g_\phi(x_i).
\]

The generator's full training objective is
\begin{equation}
  \label{eq:generator-total-objective}
  \mathcal{L}_{\mathrm{generator}}(\phi)
  =
  \mathcal{L}_{\mathrm{PG}}(\phi)
  +
  \lambda_{\mathrm{EI}}\,
  \mathcal{L}_{\mathrm{EI}}(\phi;\mathcal{B}_e),
\end{equation}
where $\lambda_{\mathrm{EI}}=1.0$ controls the strength of the reward-weighted SFT term.

\subsection{Architecture and Tokenization}
The learner and generator are independently parameterized decoder-only Llama transformers with identical architecture~\citep{touvron2023llamaopenefficientfoundation}.
We use byte-level tokenization with a fixed vocabulary of 256 byte values because the universal machine produces raw bytes, and because it enables a clean, modality-agnostic evaluation of universal prediction: how well the model predicts the next byte in arbitrary sequences encoding text, images, audio, or other data.
Programs and outputs are prefixed by the bytes \texttt{S} and \texttt{O}, respectively.
During program generation, logits are restricted to the eight Brainf*ck instructions, ten canonical single-byte macro instructions, and the end-of-program token \texttt{F}, whereas output sequences may contain any byte value.

\section{Empirical Results} 
\subsection{Universal zero-shot transfer scaling laws}
\begin{figure}[t!]
    \centering
      \includegraphics[width=1.0\linewidth]{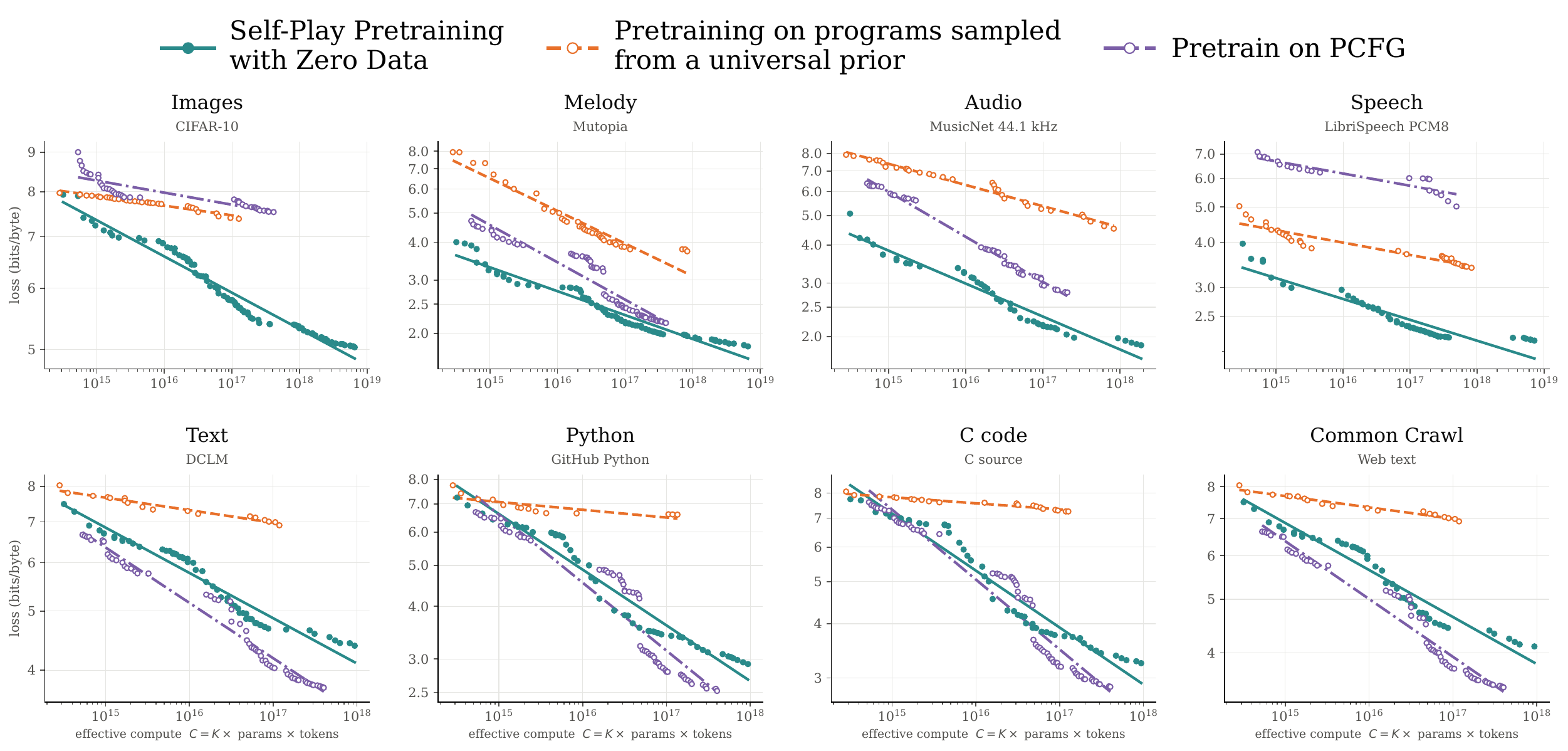}
\caption{
\textbf{Self-play learns predictive structure that transfers across modalities.}
We compare self-play to two fixed synthetic pretraining distributions:
programs sampled from a universal prior over Brainf*ck programs and probabilistic context-free grammars (PCFGs).
Sampling from the universal prior scales substantially more slowly than self-play,
showing that access to a universal space of programs alone is insufficient without the adaptive curriculum.
Pretraining on PCFG transfers strongly to language-like domains, but its benefits are
less consistent across non-language modalities.
In contrast, self-play exhibits predictable scaling across images, melody, audio, speech,
text, and code despite using minimal inductive bias.
The compute-optimal frontier ends where we do not find further models with smaller
loss than the largest-compute model shown.}
    \label{fig:randomly_sampled}
\end{figure}

\label{subsec: main-result}
In standard pretraining, scaling compute typically entails both increasing model size and training the model on more natural data.
Our scaling experiments study whether increasing compute via self-play, without any natural data, produces predictable improvements in zero-shot performance on held-out natural data.

\paragraph{Scaling recipe.}
We follow the scaling methodology of \cite{kim2026pre, wen2026fantastic}. 
At each model scale, we tune hyperparameters to \emph{local optimality}, in the sense defined in~\citet{kim2026pre}, via coordinate descent on a geometrically-spaced grid.
Because our training objective is entirely synthetic, it does not provide an obvious criterion for selecting hyperparameters; poorly tuned hyperparameters could prevent clean scaling laws. 
We therefore use validation loss averaged across DCLM and DNA as a model-selection signal. This introduces limited leakage through hyperparameter selection, but natural data are never used for gradient updates.

Because tuning every possible hyperparameter at every scale is
infeasible, we restrict the search to the most important hyperparameters based on preliminary experiments: the learner learning
rate, the generator-to-learner learning-rate \emph{ratio}, the batch size, and the
generator KL regularization coefficient $\beta$.
Due to compute constraints, we fix a large maximum training budget of 34.36B tokens rather than separately tuning the number of self-play rounds. Because we use only a short fixed warmup followed by a constant learning rate, every intermediate checkpoint is equivalent to a run stopped at that token budget. 
Thus, a single long run allows us to optimize over training duration retrospectively when constructing the scaling laws.
When tuning batch size, we hold the total number of training tokens fixed; when the batch size changes, the number of rounds is adjusted accordingly to preserve the total token budget.
Across all scales, we use a context length of 4096 tokens.
We find that ensembling across randomly initialized models is useful. 
Therefore, at the locally-optimal hyperparameters, we train $K$
independent seeds per scale and form ensembles averaging the
models' predictive distributions.

For each dataset and algorithm, we construct a \emph{compute-optimal} frontier~\citep{kaplan2020scaling} over
model sizes, checkpoints, and ensemble sizes.
A point is on the frontier if and only if it achieves
lower validation loss than every observed configuration with less than or
equal compute.

We fit each compute-optimal frontier with the asymptotic power law
\[
    L(C) = E + A C^{-\alpha},
\]
where $L$ is the observed loss in bits per byte and $E$ is a fitted
asymptotic loss floor.
We fit the model independently for each dataset.

\paragraph{Self-play exhibits universal zero-shot power-law scaling in compute.}
As shown in Figure~\ref{fig:training}, we observe scaling laws over a diverse range of modalities: text, images, and music. 
We present additional results in Figure~\ref{fig:eval_all}.
The scaling laws over these modalities are broadly similar, as seen in \Cref{tab:scaling_exponents}, (with DNA as the exceptional case).  
We discuss the implications of this in Section~\ref{sec:explaing_scaling}.
In brief, we expect this to be the case when learning universal structure rather than contingent knowledge is the bottleneck to scaling. 
Importantly, these scaling results are entirely zero-shot: that is, they arise without any gradient steps on any of the evaluation datasets. 

\paragraph{Pretraining on a fixed universal program prior exhibits slow
scaling.}
To isolate the value of self-play, we compare self-play against a non-adaptive
baseline over exactly the same program space; we use the same mixture of validation loss on DCLM and DNA.
Instead of learning a distribution over programs, the baseline samples programs
from a fixed Solomonoff-style prior~\citep{solomonoff1964formal}: instruction
tokens are drawn i.i.d.\ until termination, giving full support to every finite
program while favoring shorter descriptions.
Thus, both methods have access to the same universal space of computable
structure; they differ only in whether the sampling distribution adapts to the
learner.
Figure~\ref{fig:randomly_sampled} shows that fixed sampling scales substantially
more slowly, demonstrating that access to a universal program space alone is not
enough---self-play must learn where in that space to allocate training compute.
We offer additional evaluation datasets in~\Cref{fig:eval_all}.

Qualitatively, we see that self-play's improvement is because it discovers programs whose outputs exhibit recognizable mathematical structure~(\Cref{tab: discovered-math-structure}) far earlier than we would expect under uniform sampling from the universal prior: across $1.64 \times 10^8$ programs drawn from the uniform prior, we find no instances of any family except arithmetic sequences. Because such structures are common in mathematical modeling, this shows that our self-play algorithm can efficiently identify universal data, and that this is one mechanism driving the faster scaling observed in \Cref{fig:randomly_sampled}. 

\begin{table}[ht]
\centering
\footnotesize
\setlength{\tabcolsep}{4pt}
\begin{tabular}{lllcc}
\toprule
\makecell{\textbf{Family}\\\textbf{(mod 256)}} &
\textbf{Example program} & 
\textbf{Its output} &
\makecell{\textbf{Earliest}\\\textbf{round}} &
\makecell{$\mathbb{E}[\text{first round}]$\\\textbf{(univ.\ prior)}} \\
\midrule
Arithmetic & \texttt{S+[.++]}            & $1,3,5,7,9,\ldots$    & 0   & $\approx 105$ \\
Fibonacci  & \texttt{S,[[.C>.C>]}        & $1,1,2,3,5,\ldots$    & 512 & $>53{,}000$ \\
Geometric  & \texttt{S+[.L>]}            & $1,3,9,27,81,\ldots$  & 256 & $>53{,}000$ \\
Quadratic  & \texttt{S,.[<C>{}>VX<RX++]} & $9,25,59,111,\ldots$  & 512 & $>53{,}000$ \\
Cubic      & \texttt{S+[[-.L>L>-]-]}     & $0,254,236,74,\ldots$ & 512 & $>53{,}000$ \\
\bottomrule
\end{tabular}
\vspace{6pt}
\caption{\textbf{Program families with recognizable mathematical structure discovered by the generator during self-play.}  
\emph{Earliest round} gives the earliest round in which a member of the family first appears during training, while \emph{univ. prior} gives the expected first appearance round if programs are drawn from the universal prior, including the added primitives. See \cref{apx:emergent-math-structure} for additional details.}
\label{tab: discovered-math-structure}
\end{table}

\paragraph{PCFG pretraining is effective on language-like domains but lacks broad cross-domain transfer.}
In Figure~\ref{fig:randomly_sampled}, we also compare against pretraining on probabilistic context-free grammars
(PCFGs), which provide a hand-designed source of hierarchical and compositional
structure particularly well suited for language; for details of PCFG data generation see~\Cref{app:pcfg}.
We expect pretraining on PCFG to be highly competitive on
language-like domains, but its inductive bias is specialized to a particular
class of structure. 
In contrast, our self-play procedure is, in principle, universal. 
Consistent with this interpretation, PCFG pretraining is stronger on text and
code, where its inductive bias is well matched, while self-play substantially
outperforms it on images, music, audio, and speech. 
Thus, self-play does not always match the performance of a specialized prior on domains where that prior is particularly well suited; however, it learns
structure that transfers more broadly across modalities.
We find that models trained on PCFG and the universal prior fail on our ICL evaluations in~\Cref{fig:icl_ablation}.
\begin{figure}
    \centering
    \includegraphics[width=0.8\linewidth]{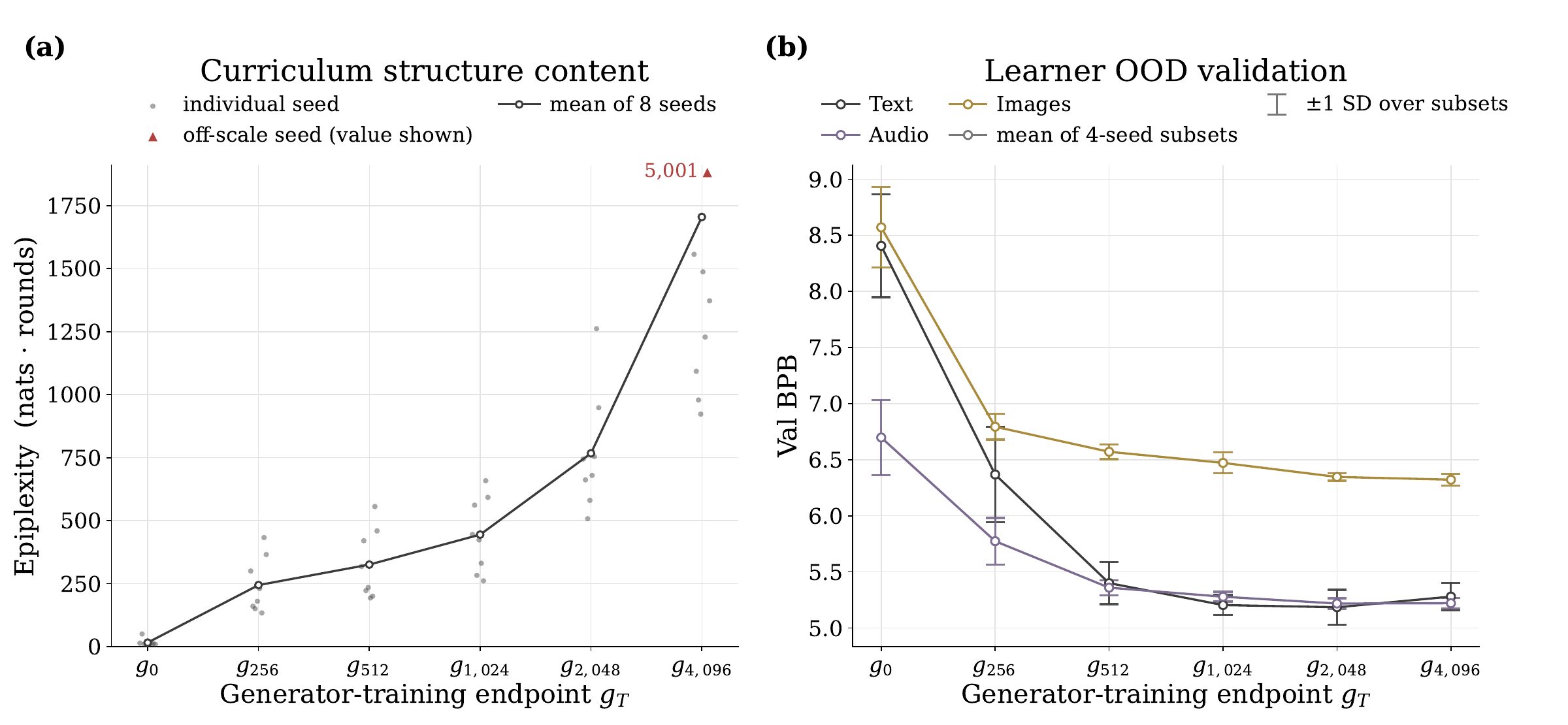}
\caption{\textbf{Later generator checkpoints provide additional, non-redundant training value.}
  For each generator-training endpoint $T>0$ we build a fixed corpus of programs by sampling from 16 generator snapshots spaced $T/16$ apart and ending at $T$; $g_0$ uses the
  untrained generator. A 1M-parameter learner is then trained from scratch on each corpus,
  four seeds per endpoint.
  \textbf{(a)} Epiplexity, the excess training loss the learner accumulates before converging,
  grows steadily with $T$: corpora written by later checkpoints contain more structure.
  \textbf{(b)} Out-of-distribution validation BPB of 4 ensembled seeds on text, audio, and images falls with $T$.
  Together, the two panels show that later checkpoints enrich the curriculum rather than
  repeating earlier material, and that this additional data improves transfer to unseen datasets.
  }
  \label{fig:curriculum-quality-vs-time}
    \label{fig:generator_value}
\end{figure}

\paragraph{The generator produces increasingly useful training data.}
We next ask whether the generator improves over the course of self-play.
For each endpoint $T$, we construct a fixed corpus $\mathcal{D}_T$ by sampling
4.19M programs uniformly across 16 generator checkpoints up to $T$ (with $g_0$
as untrained), and train a fresh 1M-parameter learner for one epoch
on each corpus under a fixed token budget.

We evaluate each corpus by its \emph{epiplexity}---the amount of structure
extractable by a compute-bounded learner~\citep{finzi2026entropyepiplexityrethinkinginformation}---and by
zero-shot performance on held-out text, audio, and images
(Figure~\ref{fig:generator_value}).
Both improve steadily with $T$: later generators produce data containing more
learnable structure and yielding better transfer, indicating that self-play
continually improves the quality of the training distribution.
Consistent with this, the generator also discovers recognizable mathematical
sequences far earlier than expected under the fixed universal prior
(\cref{tab: discovered-math-structure}).

\subsection{In Context Learning}
\begin{figure}[ht!]
    \centering    \includegraphics[width=1.0\linewidth]{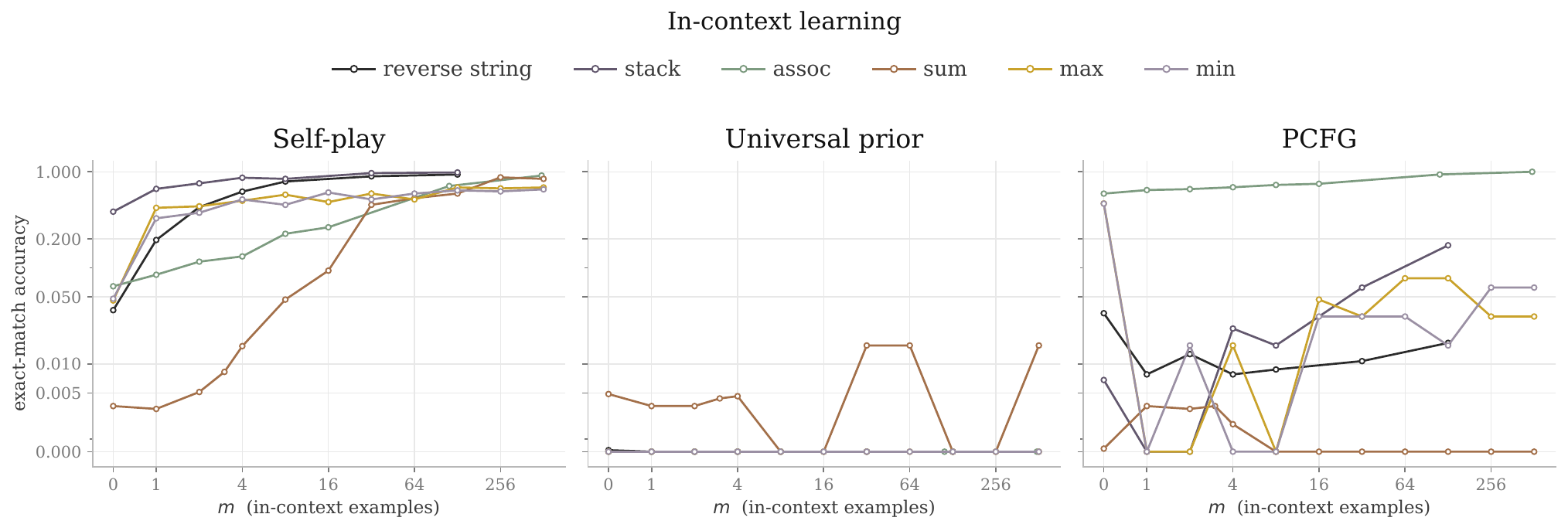}
    \caption{\textbf{Comparing ICL behavior across methods.}
    Self-play pretraining yields consistent improvements in ICL performance across all six tasks. In contrast, universal prior pretraining shows little evidence of effective ICL, while PCFG pretraining performs strongly on associative recall but transfers only weakly to the remaining tasks.
    }
    \label{fig:icl_ablation}
\end{figure}

An emergent property of large language models is their ability to infer a task from examples provided in context and apply the inferred rule to new inputs~\citep{brown2020language}.
Beyond measuring held-out loss, in-context learning provides a complementary test of whether self-play pretraining has produced a model that can infer latent structure in sequences.
We evaluate our learner's performance on several in-context learning tasks in Figure~\ref{fig:training}.
We plot the empirical success rate under greedy $\argmax$ decoding as a function of the number of in-context examples, $m$.
Importantly, the learner does not receive any additional gradient updates or fine-tuning. 
Rather, the learner must infer the latent structure underlying the sequence entirely in context.
From the perspective of universal prediction and Solomonoff induction, this is precisely the kind of behavior we would hope to emerge after our self-play procedure.
Section \ref{app:icl} contains task-specific details.

\paragraph{Self-play induces broad ICL where fixed synthetic pretraining does not.}
In Figure~\ref{fig:icl_ablation}, we see that the model can achieve almost 100\% accuracy on \textsc{reverse string} \citep{deletang2022neural}, \textsc{stack} \citep{deletang2022neural}, and \textsc{associative recall} \citep{ba2016using,graves2014neural} tasks after a sufficient number of ICL examples. The \textsc{associative recall} task shows that our model is capable of contextual search, the \textsc{reverse string} task shows that our model is capable of dynamically indexing, and \textsc{stack} shows that our model is capable of learning to simulate a context-free grammar.
In addition, we show that our model is capable of learning standard mathematical relations in-context (\textsc{max}, \textsc{min}, \textsc{sum}). This is a natural test given that our model has been trained on short programs which could naturally express the application of mathematical functions. Before any examples have been shown ($m=0$) the model already has a 6\%-8\% prediction accuracy on the \textsc{max} and \textsc{min} tasks due to its prior on copying previously produced tokens. 
In contrast, the models trained on PCFG and on samples from the universal prior over programs cannot learn all of these ICL tasks~(Figure~\ref{fig:icl_ablation}).

\paragraph{Qualitative analysis of shifting model strategies during SUM task.}
\begin{figure}[t!]
    \centering    \includegraphics[width=0.70\linewidth]{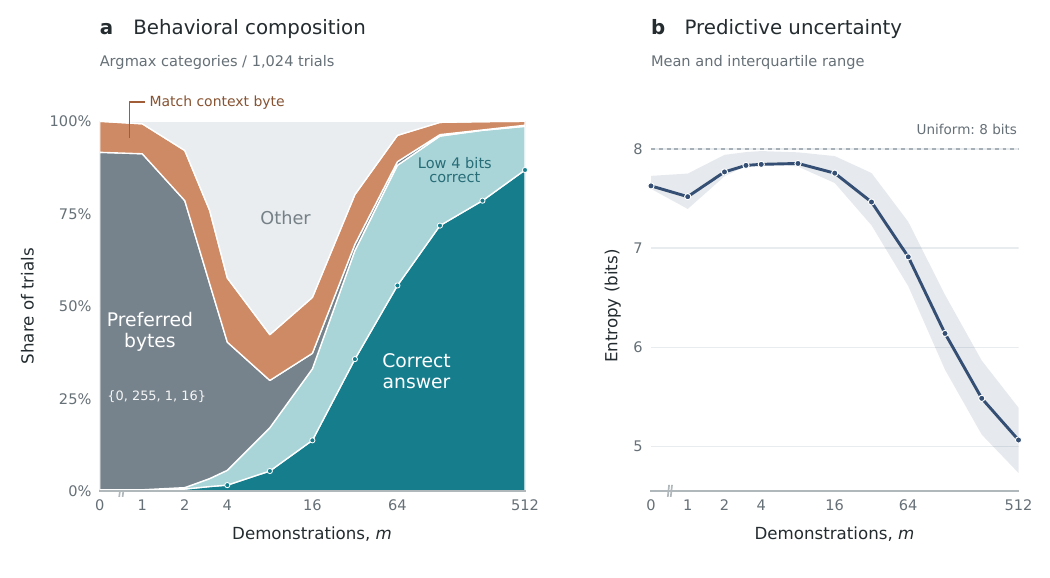}
    \caption{\textbf{Interpretable ICL behavior on \textsc{sum} task.} \textbf{(a)} Distribution over the model's inferred strategies as the number of ICL examples increases.
\textbf{(b)} Predictive entropy over the same trajectory, showing an initial loss of confidence followed by increasing certainty as the correct strategy emerges.
}    \label{fig:icl_understanding}
    
\end{figure}

In Figure~\ref{fig:icl_understanding}, we study the behavior of the model on the SUM task as it receives more ICL examples.
Initially, the model predicts trivial outputs which correspond to its prior (the marginally most common bytes, indicated in dark grey).
After seeing a few examples it begins to copy previous bytes, but then loses confidence after several overconfident but incorrect predictions, reverting to a very broad distribution; we see this reflected in the increase in entropy in the right panel.
After around 4 examples, it begins to sum the 4 low-order bits correctly and by 8 examples it begins to sum the 4 high-order bits correctly as well. From there the model improves its confidence in its strategy and \emph{locks in} on the correct approach.
\section{Explaining Self-Play Scaling laws via Universal Data Ansatz}
\label{sec:explaing_scaling}
Our experiments show that (1)  prediction on natural data improves predictably even though the learner does not train on any natural data
and (2) the observed scaling exponents are comparable to those obtained from standard pretraining on particular domains.
We propose a simple interpretation of these two results by separating two sources of predictive
information that are ordinarily entangled in natural data:
\emph{contingent information}, which is specific to the particular world or
distribution that generated the data, and \emph{universal predictive structure},
which is shared across many data-generating processes.

\paragraph{Decomposing natural-data scaling.}
\citet{hoffmann2022training} model loss as a function of model size $N$ and natural dataset size $D$ using the ansatz
\begin{equation}
L = E + \frac{A}{N^\alpha} + \frac{B}{D^\beta}. \label{eq:chinchilla_ansatz}
\end{equation}

We refine the data term by treating contingent information and universal structure as separate resources:
\begin{equation}
L = E + \frac{A}{N^\alpha}
+ \frac{B}{D_c^\beta}
+ \frac{C}{D_u^\gamma},
\label{eq:contingent_universal_loss_scaling}
\end{equation}
where $D_c$ denotes the effective amount of contingent information available to the model and $D_u$ the effective amount of universal predictive structure. 
\paragraph{Explaining self-play power law scaling in compute.}
We first use the ansatz in Equation~\ref{eq:contingent_universal_loss_scaling} to understand why we might see power law scaling in self-play compute.
Since our models are not trained on natural data,  contingent information
is fixed as training proceeds.
We absorb the third term in Equation~\ref{eq:contingent_universal_loss_scaling} into a dataset-specific constant $E'$.
Only universal predictive structure generated through self-play can grow, giving
\begin{equation}
L
=
E'
+
\frac{A}{N^\alpha}
+
\frac{C}{\left(D_u(T)\right)^\gamma}.
\end{equation}
Here we write $D_u(T)$ to indicate the amount of universal data generated via self-play at time $T$. 
If self-play generates universal structure at a power-law rate,
\begin{equation}
D_u(T) \propto T^\eta,
\end{equation}
then
\begin{equation}
L
=
E'
+
\frac{A}{N^\alpha}
+
\frac{C'}{T^{\gamma\eta}}.
\end{equation}
Thus, the ansatz directly predicts the form of the scaling law we observe:
natural-data loss can improve as a power law in the amount of self-play
training data $T$, despite no natural data being added during training, at the cost of a larger irreducible error\footnote{This must be the case for finite context lengths, but is not necessarily the case when the model has an infinite context length which it could theoretically use to learn any domain in-context.}.

\paragraph{Comparing self-play and natural-data exponents.}
Now, we show that, with some additional assumptions, the conventional one-term data-scaling
law conflates gains from learning contingent information with gains from
learning universal structure.
In standard pretraining, increasing the amount of natural data $D$ increases
both resources. 
Suppose
\begin{equation}
D_c(D) \propto D^\nu,
\qquad
D_u(D) \propto D^\mu.
\end{equation}
Substituting into \Cref{eq:contingent_universal_loss_scaling} gives
\begin{equation}
L
=
E + \frac{A}{N^\alpha}
+ \frac{B'}{D^{\beta\nu}}
+ \frac{C'}{D^{\gamma\mu}}.
\end{equation}

Asymptotically, the more slowly decaying of the two terms dominates, so we expect that a
single fitted data exponent recovers
\begin{equation}
\beta_{\mathrm{obs}}
\approx
\min\{\beta\nu,\gamma\mu\}.
\label{eq:beta_obs}
\end{equation}
We observe self-play exponents, which estimate $\gamma\eta$,
comparable to those reported for direct training on the corresponding natural
modalities (\cref{tab:scaling_exponents}).
This comparison is informative because self-play, in our setting, can improve
only by increasing universal predictive structure, whereas natural data
pretraining can improve through both universal and contingent information.
We cautiously interpret this as suggesting that learning universal structure may account for an important component of the improvements obtained by scaling
natural data.

\section{Related Work}
\label{sec:related-work}
\paragraph{Synthetic data}
Since natural data is limited, synthetic data has increasingly been explored as a promising approach.
One line of work uses generation to extract more learning signal from a fixed corpus of natural data. 
Synthetic continued pretraining generates diverse presentations and connections among facts in a small source corpus, improving the efficiency with which those facts are acquired~\citep{yang2025synthetic}. Related approaches synthesize relationships across documents, latent reasoning underlying observed text, or multiple transformations of individual documents~\citep{zelikman2024quietstarlanguagemodelsteach,ruan2025reasoninglearnlatentthoughts,kim2026dataefficientpretrainingscalingsynthetic}.
A distinct body of work asks whether useful structure can instead be learned from synthetic data that need not encode the target corpus itself. 
Pretraining on music, code, artificial languages, generic synthetic tasks, and formal languages can transfer to natural-language prediction and linguistic generalization~\citep{papadimitriou-jurafsky-2020-learning, hu2025circuitschomskyprepretrainingformal}, while analogous results show transfer from procedurally generated data to natural images~\citep{kataoka2021pretrainingnaturalimages, baradad2022learninglookingnoise}. 
These results suggest that synthetic data can teach predictive structure that is shared across domains.
Recent approaches take a dataset attribution approach to synthesizing post-training data end-to-end~\citep{thrush2026synthetic}; we view this as a complementary approach for pretraining data.

\paragraph{Self-play}
Early work on intrinsic motivation proposed rewarding agents for learning or compression progress, thereby directing exploration toward regularities that are neither already mastered nor currently unlearnable \citep{schmidhuber2009drivencompressionprogresssimple}. Related work on automatic curriculum and open-ended learning similarly constructs tasks that remain near the learner’s frontier \citep{schmidhuber2012powerplaytrainingincreasinglygeneral}.
More recently, self-play has been applied to language models in formal math~\citep{poesia2024learningformalmathematicsintrinsic, bailey2026scalingselfplayselfguidance,dong2025stpselfplayllmtheorem,dong2024formaltheoremprovingrewarding}, reasoning~\citep{zhao2025absolutezeroreinforcedselfplay, liu2026spadeselfplayadaptivesynthetic, chen2025selfquestioninglanguagemodels}, coding~\citep{wang2026codea1adversarialevolvingcode,choi2026anchoredselfplaycoderepair}, and agentic tool use~\citep{zhou2025selfchallenginglanguagemodelagents,acikgoz2026toolr0selfevolvingllmagents}.
In contrast, we focus on pretraining with the scientific goal of establishing whether self-play can be used to discover training distributions whose structure transfers to prediction on completely unseen natural data.
Moreover, for a clean analysis, our generator and learner begin from scratch, while most previous work, with the exception of ~\citet{poesia2024learningformalmathematicsintrinsic}, uses pretrained language models.

\paragraph{Universal prediction and algorithmic pretraining}
Universal prediction provides a theoretical framework for understanding when prediction over arbitrary computable data-generating processes is possible \citep{solomonoff1964formal,merhav98}. Most relevant to our work are \citet{graumoya2024learninguniversalpredictors} and ~\citet{bloem2025universalpretrainingiteratedrandom}.
~\citet{graumoya2024learninguniversalpredictors} show that neural networks can amortize universal prediction by training on outputs of programs sampled from a universal Turing machine, and demonstrate transfer to held-out algorithmic processes. 
We instead learn the program distribution jointly with the learner, rewarding programs according to the learning progress they induce. 
Closest empirically to our setting, \citet{bloem2025universalpretrainingiteratedrandom} provides evidence for \emph{universal pretraining} from zero natural data, showing that training on sequences produced by iterated random computation yields improving zero-shot prediction on unseen real-world data with model scale and can accelerate subsequent finetuning.
Our work builds on this direction by making the data-generating distribution adaptive: we formulate program generation as an RL problem and reward programs according to the learning progress they induce in the learner.
This yields a learned curriculum that co-evolves with the learner.
Empirically, we demonstrate transfer on a broader suite of held-out natural datasets and explicitly fit and analyze scaling laws for performance under self-play pretraining.

\paragraph{Epiplexity}
\cite{finzi2026entropyepiplexityrethinkinginformation} formalize a measure of the amount of structured information in data relative to a compute-bounded observer, which is necessary because unbounded notions such as entropy and Kolmogorov complexity fail to capture emergent structure. For example, for AlphaZero, the Kolmogorov complexity of its output is upper bounded by the length of the program that generated it, while the resulting model is vastly larger, since it stores what that program learned along the way in the weights of a neural network, amortizing inference. Just as Go can be decided with brute-force search, our setting can be trivially solved by an unbounded-compute agent via Solomonoff induction; only at finite compute does searching for useful recurring patterns ahead of time become necessary. We use epiplexity as a measure to check that our generator's output increases in quality (see \cref{fig:curriculum-quality-vs-time}). Furthermore, such a notion of structure offers a potential theoretical basis for the design of universal reward functions.
\section{Discussion}
Our results show that predictable improvements in next-token prediction on natural data can emerge even when no natural data is used for training.
This provides evidence that some of the structure ordinarily acquired through standard pretraining on natural data is beneficial because it teaches the model universal predictive regularities.
At the same time, universal pretraining cannot recover contingent information: facts about a particular world must ultimately enter through interaction with that world.
Therefore, we do not view universal pretraining as a replacement for natural data, but as a way to isolate universal structure and study whether that component can instead be generated from compute.

\paragraph{Implications on synthetic data.}
The decomposition in~\Cref{eq:contingent_universal_loss_scaling} provides a useful way to interpret recent approaches to synthetic data.
Some methods generate abstract, formal, procedural, or otherwise domain-independent data whose primary value is to expose the model to transferable structure
\citep{hu2025circuitschomskyprepretrainingformal,papadimitriou-jurafsky-2020-learning}.
These methods can be understood as increasing the effective supply of universal data, $D_u$.
Other methods begin with a fixed natural corpus and generate new presentations~\citep{yang2025synthetic}, relations, or latent reasoning traces from it
\citep{kim2026dataefficientpretrainingscalingsynthetic,ruan2025reasoninglearnlatentthoughts, zelikman2024quietstarlanguagemodelsteach}.
Such methods can instead improve the efficiency with which information already present in the natural corpus is acquired, effectively increasing the useful supply of $D_c$; in practice, the chain-of-thought based methods may also expose additional reusable structure and therefore also affect $\mathcal{D}_u$.
This perspective offers an explanation for why synthetic data can improve pretraining even when the data is not from the same distribution as the target distribution of natural language.
If universal structure is a limiting resource, then generating additional structured experience can improve prediction despite bearing little surface resemblance to the target distribution.
If contingent information is limiting, synthetic transformations of a fixed corpus can increase the amount of learning signal extracted from each observation.
Our results demonstrate an extreme point in this design space: $D_c$ is held fixed at zero during training, while the learning system is allowed to spend increasing compute searching for $D_u$.

\paragraph{Why {\textit{tabula rasa?}}}
Our {\textit{tabula rasa}} setting is intended as a controlled scientific experiment rather than necessarily the most practical way to pretrain a model.
Since the learner and generator are randomly initialized, any transfer to natural data must have been acquired through the self-play process.
In practice, there is no requirement that self-play begin from scratch.
The key question is whether self-generated experience can continue expanding the frontier after naturally available data has become expensive, redundant, or exhausted.
Our results suggest that this possibility is worth studying: if an adaptive curriculum can bootstrap transferable structure from random initialization, then the same mechanism may also be useful when initialized from a non-random learner.
This could make our self-play algorithm complementary to standard pretraining.

\paragraph{Future Directions}
The experiments reported here are confined to models below 25M parameters at a 4K context, and the most immediate question is whether these results persist for larger models. Achieving this may require a more expressive programming language, allowing reusable abstractions to co-evolve with the generator, alongside other modifications that improve program search efficiency and overall scalability. 
Future work could also test whether the discovered mathematical structures causally contribute to transfer through circuit analysis or curriculum ablations.

\section{Acknowledgments}
We especially thank Suhas Kotha and Marvin Li for detailed feedback on an earlier draft of the paper. We thank Xiao-Liang Qi for valuable discussions at the inception of this project. Kfir Dolev was supported by the Long Term Future Fund and later by the Zuckerman STEM leadership program.

\bibliographystyle{plainnat}
\bibliography{iclr2027_conference}
\appendix
\section{Additional experimental results}
\begin{table}[t]
  \centering
  \small
  \begin{tabular}{l r r l}
  \toprule
  Modality / dataset & Ours $b$ & Literature $b$ & Ref. \\
  \midrule
  \multicolumn{4}{l}{\emph{Text}} \\
  text (dclm) & $0.123$ & $0.048$--$0.099^{\dagger}$ & \citet{henighan2020scaling,aghajanyan2023scaling} \\
  \midrule
  \multicolumn{4}{l}{\emph{Images}} \\
  CIFAR-10 (HWC interl.) & $0.066$ &  &  \\
  CIFAR-10 image bytes   & $0.145$ & $0.065^{\dagger}$--$0.10$ & \citet{aghajanyan2023scaling}; \citet{henighan2020scaling} \\
  \midrule
  \multicolumn{4}{l}{\emph{Audio / speech}} \\
  audio 16-bit PCM               & $0.141$ &  &  \\
  audio 8-bit PCM                & $0.260$ & $0.12^{\dagger}$--$0.14^{\dagger}$ & \citet{cuervo2024scaling,aghajanyan2023scaling} \\
  MIDI (Mutopia, 16th note grid) & $0.249$ & -- &  \\
  \midrule
  \multicolumn{4}{l}{\emph{Math / formal}} \\
  Metamath set.mm & $0.129$ & $0.17$ & \citet{henighan2020scaling} \\
  \midrule
  \multicolumn{4}{l}{\emph{Biological sequences}} \\
  DNA (8-symbol) & $0.435$ & $0.01$--$0.06$ & \citet{shah2026dnahnet} \\
  \midrule
  \multicolumn{4}{l}{\emph{Code}} \\
  AITDCC C source        & $0.116$ & $0.17^{\dagger}$ & \citet{aghajanyan2023scaling} \\
  Python source (GitHub) & $0.113$ &  &  \\
  \bottomrule
  \end{tabular}
  \caption{
Per-modality compute exponents $b$ from fits $L(C)=A\,C^{-b}+E$. Values marked $^{\dagger}$ are derived from published Chinchilla-form fits via $b=\alpha\beta/(\alpha+\beta)$ under the compute-optimal allocation. Where a range is given, values correspond to the citations in order. Dashes indicate that the authors could not find a published exponent for that modality. Exponents from \method{} are broadly similar to exponents from pre-training on the literature, if a bit higher. A detailed description and an illustration of the datasets can be found in \cref{app:benchmarks}}
  \label{tab:scaling_exponents}
  \end{table}

\subsection{Pre-Pretraining with Self-Play Accelerates Pretraining on Natural Data}
\label{sec:pre_pretraining}

Our scaling results show that self-play produces transferable structure; a
natural follow-up question is whether that structure remains useful once
natural data becomes available. We therefore treat self-play as
\emph{pre-pretraining}~\citep{hu2025circuitschomskyprepretrainingformal}: a learner produced by self-play is used as the
initialization for ordinary pretraining on natural data, and we compare it
against the same architecture pretrained from random initialization.

\begin{figure}[t]
\centering
\includegraphics[width=\linewidth]{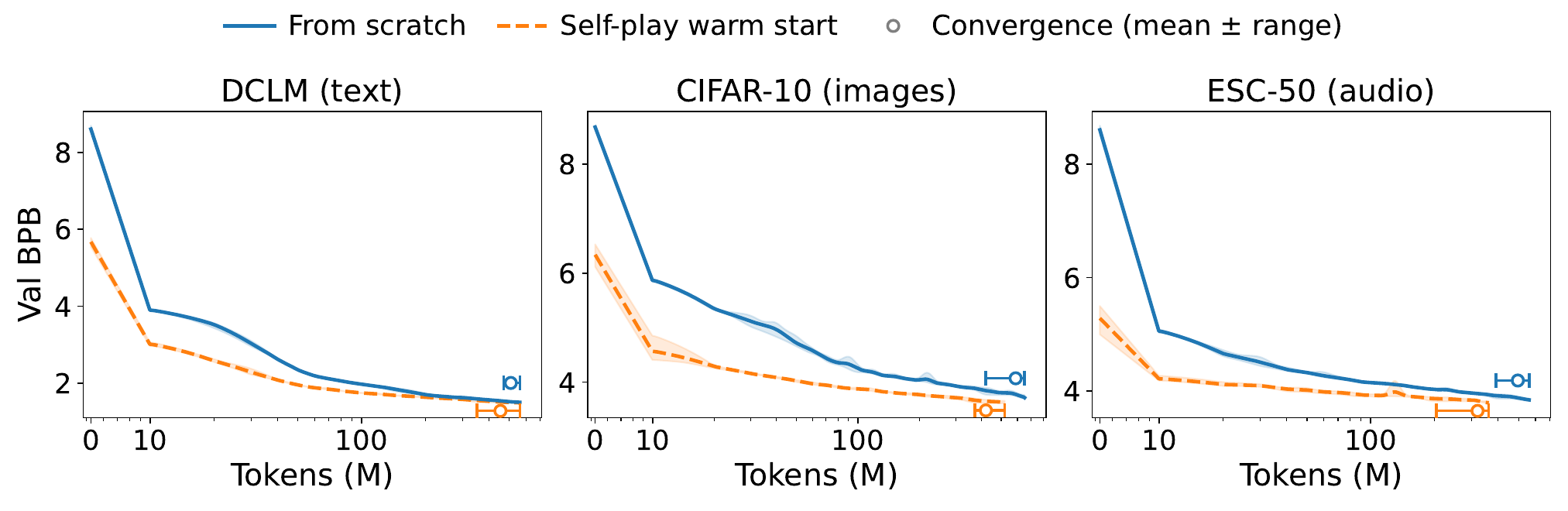}
\caption{\textbf{Self-play pre-pretraining accelerates pretraining on natural
data.} Validation BPB during pretraining of a 24.4M model on three
byte-encoded modalities, from random initialization (blue) and from the final
self-play checkpoint (orange), at each arm's best hyperparameters (mean over
4 seeds; shaded band spans the seed range). The leftmost point (marked $0$)
is the validation loss before any natural-data training --- for the warm
start, its zero-shot transfer --- and the token axis is logarithmic from the
first evaluation onward. Open markers show the mean tokens consumed at
convergence, with horizontal bars spanning the seed range (markers vertically
offset for legibility); Throughout training, the warm start reaches
every loss level first.
}
\label{fig:pre_pretraining}
\end{figure}

\paragraph{Setup.}
We compare downstream pretraining of a 24.4M-parameter model from two
initializations: random weights (``from scratch'') and the final self-play
checkpoint (``self-play warm start'').
We evaluate both on DCLM text, CIFAR-10 images, and ESC-50 audio, using the
same fixed natural-data corpus for each modality.
Both methods are trained to convergence, with learning rate and weight decay tuned separately for each.

\paragraph{Results}
Figure~\ref{fig:pre_pretraining} shows that self-play pre-pretraining accelerates downstream training across all three modalities. 
The warm-start models start with a lower loss than random initialization, as expected, and retain an advantage throughout training, reaching a low validation-loss level with fewer natural-data tokens. 
The savings are substantial on ESC-50 ($320$M vs.\ $496$M tokens) and CIFAR-10 ($421$M vs.\ $588$M). By the end of our training protocol, the loss gap has narrowed considerably, indicating that the clearest benefit of self-play pre-pretraining is accelerated learning from natural data.
\begin{figure}[ht!]
    \centering
      \includegraphics[width=0.8\linewidth]{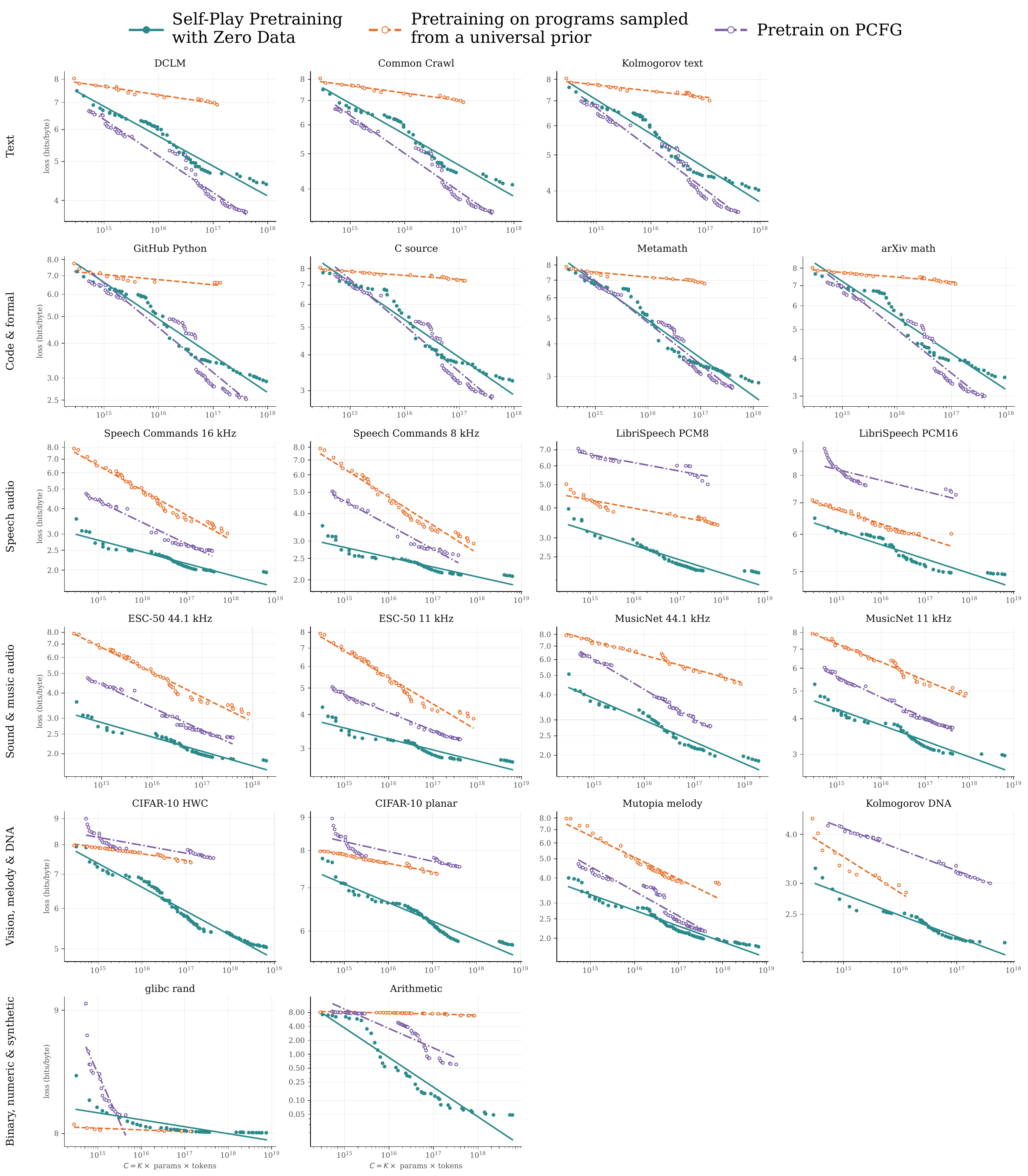}
\caption{
\textbf{Scaling laws across all evaluation datasets.}
}
    \label{fig:eval_all}
\end{figure}


\subsection{Pre-pretraining details}
\label{app:pre_pretraining}
After the validation split, the training corpora contain approximately 255M
(DCLM), 146M (CIFAR-10), and 152M (ESC-50) tokens, and runs repeat this data
over epochs, terminating at approximate convergence.  The learning rate is
held constant until validation BPB plateaus (improvement ${<}\,0.005$ for 5
consecutive evaluations), then decayed to zero over a 200-step cosine
schedule; we report the \emph{converged BPB}, the validation loss after this
final decay. Learning rate and weight decay are tuned separately for each arm,
from $\mathrm{LR}\in\{10^{-3},\,3\!\times\!10^{-3},\,10^{-2}\}$ and
$\mathrm{WD}\in\{0.0,\,0.1,\,0.3,\,0.8\}$ with 4 seeds per configuration; per
arm we select the configuration with the lowest mean converged BPB across
seeds.
We do not count the compute spent on self-play pre-pretraining itself in this
comparison: it is a one-time cost that amortizes across downstream training
runs --- here, a single self-play checkpoint initializes the warm arm on all
three modalities --- in the same way that a pretrained checkpoint is reused
across many fine-tuning tasks.

\section{Benchmark details}\label{app:benchmarks}

Here we collect more details on the benchmarks. To test the ability of the model to predict intrinsically different kinds of "natural" sequences, we designed a diverse set of benchmarks generated by different processes. Regardless of the provenance, they share the same interface, obtained by encoding them as byte sequences.

\subsection{Natural text}

To construct this benchmark, we used DCLM-Baseline-1.0, a filtered collection of web text extracted from Common Crawl.  Explicitly, we sampled the ten global DCLM shards so that the benchmark did
not come from only one part of the collection. Each DCLM record is stored in a JSON container, but the benchmark retains only
its text field.  We encode this text directly as UTF-8 bytes, concatenate the
document texts within each shard, and divide the result into fixed windows for
next-byte prediction.  The predictor therefore has to exploit regularities such as spelling,
punctuation, word structure, and local syntax through the same byte interface
used for every other benchmark.

\begin{figure}[htbp]
  \centering
  \includegraphics[width=\linewidth]
    {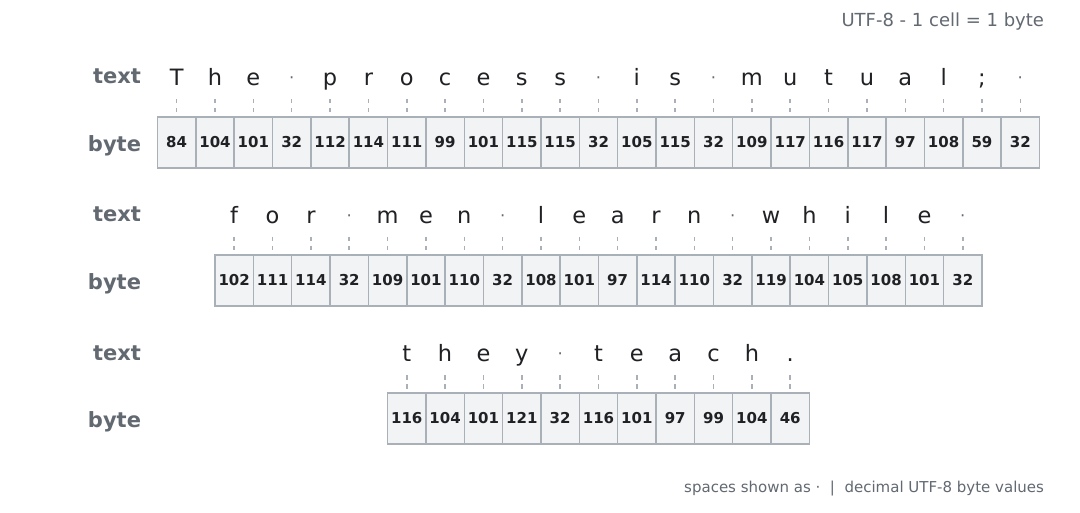}
  \caption{Illustration of the natural-text encoding for an illustrative ASCII
    excerpt. Each cell shows one decimal UTF-8 byte; middle dots mark spaces
    (byte 32).}
  \label{fig:dclm-text-encoding}
\end{figure}

\subsection{Natural images}

To construct this benchmark, we used the official CIFAR-10 test batch.  It
contains $32\times32$ color images, each paired with one of ten class labels.
Since our task is next-byte prediction rather than classification, we remove the labels and retain only the unsigned 8-bit RGB pixel values.  We also exclude filenames, image headers, compression, and other metadata, so the predictor sees the common byte interface.

A two-dimensional image must still be arranged as a one-dimensional byte sequence.  We included two lossless encodings of exactly the same images.  The
planar (CHW) encoding traverses each image row by row, storing all red values, then all green, and then all blue.  The interleaved (HWC) encoding uses the same pixel order but places each pixel's R, G, and B values next to one another.
We considered both encodings to check whether this one dimensional ordering made
a difference to next-byte prediction performance, but we did not find any appreciable effect.
\begin{figure}[htbp]
  \centering
  \includegraphics[width=\linewidth]
    {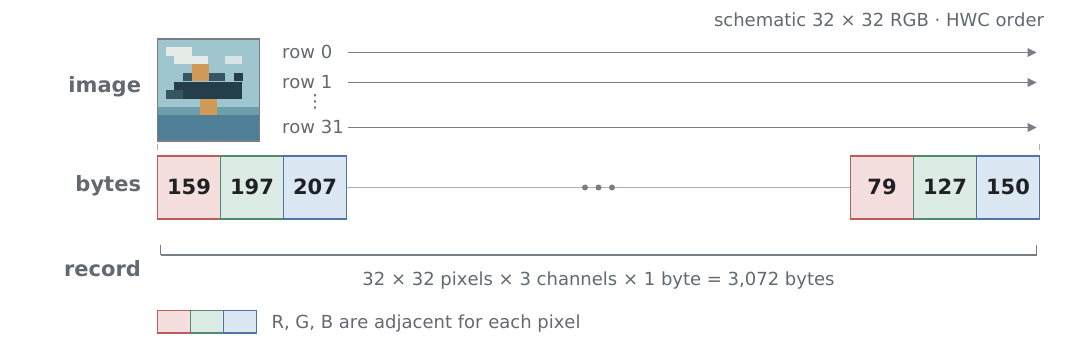}
  \caption{Illustration of the interleaved (HWC) encoding, in which the RGB
    values of each pixel are adjacent. The thumbnail and displayed byte values
    are schematic.}
  \label{fig:cifar10-hwc-encoding}
\end{figure}

\subsection{RAW audio: Natural speech}

Our raw-audio benchmarks use the official Speech Commands v0.02 test archive, released under CC BY 4.0.  It contains 4,890 one-second mono recordings, encoded in WAV files, composed by ten target commands together with unknown-word and silence examples.

A WAV file mixes the waveform with a header, and its samples are signed 16-bit little-endian values.  We discard the header, paths, category labels, and other
metadata.  Each amplitude $s$ is clipped to the PCM16 range and mapped to the
unsigned byte \(\lfloor(s+32768)/256\rfloor\), so zero amplitude becomes 128 and
neighboring bytes remain neighboring points in time. 

We provide three versions of the same recordings.  The 16\,kHz version quantizes the original samples directly.  For the 8- and 4\,kHz versions, fixed anti-aliasing filters downsample each recording independently by factors of two and four before the same quantization.  These benchmarks test short-time acoustic prediction. With context length 256, each
record contains 255 consecutive waveform bytes. The 255 bytes span 15.94\,ms at 16\,kHz, 31.88\,ms at
8\,kHz, and 63.75\,ms at 4\,kHz.
Reducing the sample
rate sacrifices high-frequency detail but exposes a longer interval to the same bounded-context model.

\subsection{Music}

Raw audio is physically direct but temporally expensive.  PCM8 sampled at
16\,kHz consumes 16,000 bytes per second; even the repository's 4\,kHz PCM8
variant consumes 4,000 bytes per second.  So even a 4096 context length is only able to understand local features. Symbolic score music representation is instead much more compact. At four bytes per quarter note, 255 bytes represent 63.75 quarter-note grid units.  They may contain several phrases or a
substantial portion of a movement rather than a fraction of one acoustic event.

To construct this benchmark, we used the Mutopia project data. The Mutopia Project is a volunteer collection of open sheet music written in
LilyPond and based on editions in the public domain
\cite{mutopiarepo}.  Its contribution pages provide downloadable notation,
PDF, and MIDI artifacts together with source, maintainer, typesetting, and
license . 

From the full Mutopia project, we took a selection of 40 highly recognizable Western classical pieces under Public Domain license, including familiar pieces by Beethoven, Mozart, Bach,  Chopin, Debussy, Schubert, Tchaikovsky, and others. 
A Standard MIDI File serializes a technical event stream rather than a direct
sequence of musical states.  It can contain a header,
format and track declarations, variable-length delta encodings, tempo events,
time signatures, program changes, channels, velocities, controllers, text,
copyright notices, names, and end-of-track events.  Two files representing
substantially the same score can differ in many raw bytes because of exporter,
track layout, event ordering, and metadata choices.
We stripped all this data, retaining only the melodies. 

To convert the complex music information in the MIDI melodies into a simple "time-sequence" byte encoding, we quantized the scores in 16th notes, each byte in the benchmark corresponding the state of that time grid cell: either a new note, a continuation of the previous one, or a silence. The selected track need not be monophonic.  The benchmark creates a
monophonic output by choosing at most one active source note in every grid
cell.

We use byte value 0-127 to encode the absolute MIDI pitch . MIDI pitch is an absolute semitone number: 60 is middle C, 61 is C-sharp/D-flat, 62 is D, 63 is D-sharp/E-flat, 64 is E, and 67 is G. Byte value 128 represents continuing holding the previous note in the current byte cell, 129 represents a pause, and 130 denotes the end of a piece. In this benchmark, the bytes 131-255 are unused.

\begin{figure}[htbp]
  \centering
  \includegraphics[width=\linewidth]
    {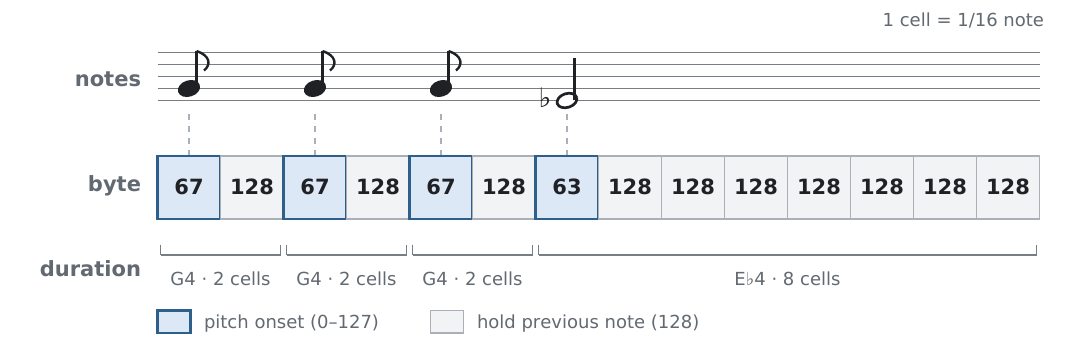}
  \caption{Qualitative illustration of the melody byte encoding using the
    opening motif of Beethoven's Fifth Symphony. Each cell is one sixteenth
    note: bytes 0--127 start a pitch, while byte 128 holds the previous note.}
  \label{fig:midi-encoding}
\end{figure}

Together with the benchmark data, we provide MIDI files reconstructed from this simplified byte encoding, which we used to check the main themes were still recognizable. 

We truncated the longer pieces to 4095 bytes.

\subsection{DNA}
\label{sub:kolmogorov-dna}

To construct this benchmark, we used the large DNA dataset released with the
KoLMogorov Test \cite{yoran2025kolmogorov}.  The original
test asks code-generating models to produce short programs that reproduce a
sequence exactly.  Here we reuse only its released DNA sequences for ordinary
next-symbol prediction, and refer to the resulting benchmark as
\texttt{kolmogorov\_dna}. 

The KoLMogorov paper describes this stream as derived from GRCh38, the curated
human reference assembly \cite{ncbiGRCh38}.  A reference assembly is a
composite sequence assembled and maintained as a common coordinate system; it
is not a file of raw sequencing reads or the observed genome of one person.
The source retains upper- and lower-case versions of the four bases A, C, G,
and T as eight distinct symbols.  Lower-case sequence is commonly used for
soft masking, which marks repetitive or low-complexity regions while retaining
the underlying base.
The released \code{dna.bin} contains the numeric
values 0-7, using the encoding
\[
0\mapsto\mathtt{a},\quad 1\mapsto\mathtt{c},\quad
2\mapsto\mathtt{t},\quad 3\mapsto\mathtt{g},\quad
4\mapsto\mathtt{A},\quad 5\mapsto\mathtt{C},\quad
6\mapsto\mathtt{T},\quad 7\mapsto\mathtt{G}.
\]
From the large release, we retain its first
\(32\,\mathrm{MiB}\) numeric symbols.  We divide this prefix
without overlap into 131,586 consecutive records of 255 symbols, discarding the final tail.
A predictor told that only eight values are possible could obtain 3 BPB by
assigning them equal probability, whereas a uniform prediction over the
learner's full 256-byte output space costs 8 BPB.  The learner must therefore
recognize the compact alphabet as well as exploit local base composition,
masking runs, repeats, and motifs visible within 255 symbols.

\subsection{Formal mathematics: Metamath}
\label{sub:metamath}

To construct this benchmark, we used a fixed revision of \texttt{set.mm}, the
main Metamath database \cite{setmmRepository}.  Metamath provides a simple
language for writing and checking mathematical proofs.  The database contains
declarations of symbols, hypotheses, axioms, and theorem statements with their
proofs.  Proofs refer to earlier statements by their labels and are stored in
a compressed form, using compact character strings to encode the proof steps.

We remove the comments delimited by \texttt{\$(} and \texttt{\$)}, including
explanatory prose, and discard empty lines.  Within each remaining line, we
replace runs of whitespace by a single space, remove leading and trailing
whitespace, and end the line with one newline byte.  We retain the formal
content in its source order, including the statement labels, formulas, and
compressed proofs.  The resulting text is represented directly by its ASCII
byte values, with spaces and newlines included in the sequence.

With context length 256, we divide this stream without overlap into 143,166
consecutive records of 255 bytes, discarding the final 176 bytes.  The windows
can cross line, statement, and proof boundaries.  The predictor therefore has
to exploit regularities such as recurring syntax, formula fragments, labels,
and patterns in the compressed proofs through the same byte interface used
for every other benchmark.  This tests next-byte prediction of formal
mathematical text; the model is not asked to construct or verify a proof.
\subsection{C source code}
\label{sub:aitdcc-c-source}

To construct this benchmark, we used file B from the Algorithmic Information
Theory Data Compression Challenge (AITDCC) \cite{aitdccCSource}.  The challenge
compares lossless compressors across different kinds of data.  Here we reuse
its released C source for next-byte prediction.  We extract the complete file
from a fixed revision of the official AITDCC archive; it contains 1,168,767
bytes of source text.

The source is ASCII text, and we retain its original byte values.  Comments,
copyright notices, preprocessor directives, identifiers, and literals remain,
together with indentation, tabs, spaces, blank lines, and newlines.  We do not
parse or compile the source, normalize its formatting, or use a C-specific
tokenizer.  The predictor receives the original text as one continuous byte
stream.

With context length 256, we divide this stream without overlap into 4,583
consecutive records of 255 bytes, discarding the final 102 bytes.  The windows
follow byte positions and can cross line, statement, function, and source-file
boundaries.  The predictor therefore has to exploit regularities such as C
syntax, recurring identifiers, comments, and formatting through the same byte
interface used for every other benchmark.  Repeated names and code patterns
provide structure beyond individual characters, while comments also retain
the spelling and local syntax of natural language.

\section{Emergent Mathematical Structure}
\label{apx:emergent-math-structure}

During self-play, the generator discovers programs whose output tapes exhibit
recognizable mathematical structure. We search the generated programs from our
model-scaling experiments for five families of sequences: arithmetic,
quadratic, and cubic sequences; Fibonacci-like sequences; and geometric
sequences. Generated programs from the self-play runs were retained only once
every 256 training rounds, so discovery times can be measured only at this
256-round resolution. In contrast, the uniform-sampling baseline described
below checks programs at every round. Thus the performance gap between self-play and the random baseline is likely even wider, as the reported self-play discovery rounds
are therefore conservative upper bounds on the true first occurrence of each
structure. Table~\ref{tab:discovered-math-structure} summarizes the observed
families and compares their discovery times with uniform program sampling.

\paragraph{Detection criteria.}
We allow up to 30 unrelated leading bytes before the structured portion of a
tape begins. The remainder of the tape must satisfy the corresponding
recurrence modulo 256. Arithmetic, quadratic, and cubic sequences are defined
by constant first, second, and third finite differences, respectively, with a
sequence assigned to the lowest-order family it satisfies. Fibonacci-like
sequences satisfy
\[
    v_n = v_{n-1} + v_{n-2} \pmod{256},
\]
for an arbitrary seed pair, while geometric sequences satisfy
\[
    v_n = r v_{n-1} \pmod{256}
\]
for some integer ratio \(r\).

To eliminate degenerate matches, we additionally require a
minimal period of at least 30, evaluated both over the full matched region and
over its trailing window. This excludes, for example, sequences with a
structured transient followed by a constant tail.

\begin{table}[ht]
\centering
\footnotesize
\setlength{\tabcolsep}{4pt}
\begin{tabular}{lllcc}
\toprule
\makecell{\textbf{Family}\\\textbf{(mod 256)}} &
\textbf{Example program} & 
\textbf{Its output} &
\makecell{\textbf{Earliest}\\\textbf{round}} &
\makecell{$\mathbb{E}[\text{first round}]$\\\textbf{(univ.\ prior)}} \\
\midrule
Arithmetic & \texttt{S+[.++]}            & $1,3,5,7,9,\ldots$    & 0   & $\approx 105$ \\
Fibonacci  & \texttt{S,[[.C>.C>]}        & $1,1,2,3,5,\ldots$    & 512 & $>53{,}000$ \\
Geometric  & \texttt{S+[.L>]}            & $1,3,9,27,81,\ldots$  & 256 & $>53{,}000$ \\
Quadratic  & \texttt{S,.[<C>{}>VX<RX++]} & $9,25,59,111,\ldots$  & 512 & $>53{,}000$ \\
Cubic      & \texttt{S+[[-.L>L>-]-]}     & $0,254,236,74,\ldots$ & 512 & $>53{,}000$ \\
\bottomrule
\end{tabular}
\caption{Mathematical sequence families discovered in our scaling experiments.
Self-play programs were retained once every 256 rounds, so the reported
discovery rounds are the earliest saved rounds at which each family was
observed.}
\label{tab:discovered-math-structure}
\end{table}

\paragraph{Comparison with uniform program sampling.}
To estimate how readily the same structures would be discovered without
self-play, we sample programs by uniformly sampling the primitive augmented alphabet until an ``F" symbol is drawn. We draw \(1.64\times10^8\) programs in
total. A sample is counted as a hit whenever its output satisfies the same
family-level detection criterion above; it need not match the example program
shown in Table~\ref{tab:discovered-math-structure}.

For comparison with the scaling experiments, we group samples at the same
rate of 1,024 newly generated programs per round. Unlike the self-play
analysis, however, the uniform baseline is checked at every round rather than
only once every 256 rounds. The comparison therefore underestimates the gap.

The arithmetic family occurs 1,526 times in the uniform baseline,
giving an estimated probability of 
\(9.3\times10^{-6}\) and an expected first-discovery round of
approximately 105 (95\% CI: 100--111).

We observe no Fibonacci, geometric, quadratic, or cubic matches in the
\(1.64\times10^8\) baseline samples. By the rule of three, this gives the
one-sided 95\% bound
\[
    p \leq 1.8\times10^{-8},
\]
corresponding to an expected first-discovery round greater than 53,000 at
1,024 programs per round. Despite the coarser observation schedule for
self-play, all four families are observed there by round 512, compared with
no occurrences in more than 53,000 rounds' worth of uniformly sampled
programs. Because all four families have zero observed baseline hits, however,
the sampling experiment establishes only a common lower bound on their rarity
and does not determine their relative frequencies.

\section{ICL Tasks}\label{app:icl}
Each ICL task has the form
\begin{equation}
    \label{eq:icl_format}
    \big[0, x^1_1, \dots, x^1_k, f(x^1_{1\dots k})\big], \dots,
    \big[0, x^m_1, \dots, x^m_k, f(x^m_{1\dots k})\big], 
    \big[0, x^{m+1}_1, x^{m+1}_2, \dots, x^{m+1}_k, ~\bullet ~].
\end{equation}
Each example is prepended by a sentinel '0' byte and is followed by the $k$ inputs bytes and then the function applied to those bytes (brackets are for visual separation and do not appear in the actual byte sequence). After $m$ examples are shown the next example sentinel and input are passed to the model which fills in its prediction. The score for the model is the probability that it predicts the correct token, $f(x^{m+1}_{1\dots k})$.

We take $f$ to range over the tasks
\begin{enumerate}
    \item \textsc{reverse string}: Given a word $x_1x_2\cdots x_k$ reverse it. $f(x_1,x_2\dots x_k) = (x_k, x_{k-1}, dots x_1)$
    \item Stack: given a series of stack operations return the result after a final pop. Stack operations are either push, or pop, sampled with equal probabilities when the stack height is at least one. Push and pop are represented by the bytes 250 and 251 respectively. The byte after is either the argument for push, or the result for pop. Example $250, 1, 250, 2, 251, 2, 250, 3, 251$ would be followed by $3$. 
    \item \textsc{associative recall} is a dictionary association task. The dictionary has size $V$ and maps bytes from $1-255$ to bytes from $1 - 255$ (repetition allowed). The dictionary is first printed in the form of key-value pairs in the format of \cref{eq:icl_format} so that all pairs are visible to the model. Evaluation proceeds in a similar manner, with random key-value pairs sampled and presented to the model. Pairs are drawn at random after the initial print and do repeat.
    \item \textsc{sum} is the task of summing two bytes \textit{mod} $256$. The format is $[0, x_1, x_2, (x_1 + x_2) \textit{ mod } 256]$. $x_1, x_2 \neq 0$.
    \item \textsc{max} / \textsc{min} are the task of finding the min and max over the $k$ input bytes. That is $f(x_1, \dots x_k) = \operatorname{min}(x_1, \dots x_k)$ and similarly for max.
\end{enumerate}

\section{Brainf*ck details}
\label{apx:turing-machine}
The generator produces programs in a minimal Turing complete language.
We use a Brainf*ck\footnote{For completeness, this stands for Brainfuck.}-like
universal machine, similar to the variant introduced in ~\cite{graumoya2024learninguniversalpredictors}, whose eight single-character instructions move the head
between cells (\texttt{<}, \texttt{>}), increment or decrement the cell under
it (\texttt{+}, \texttt{-}), loop (\texttt{[}, \texttt{]}), and read or write
bytes (\texttt{,} and \texttt{.}).

Programs are strings over the alphabet
$\mathcal{A} = \{\texttt{<}, \texttt{>}, \texttt{+}, \texttt{-}, \texttt{[},
\texttt{]}, \texttt{.}, \texttt{,}, \texttt{F}\}$
of this machine $U$, where \texttt{F} is an
explicit end-of-program token. A program $x \in \mathcal{A}^{\le L}$ is executed
on $U$ under a bounded step and memory budget, with tape cells taken modulo a
fixed modulus $m=256$; the input instruction (\texttt{,}) reads i.i.d.\ uniform random
bytes from a random tape $\omega$, so execution defines an output distribution per program. The output
\[
  y \;=\; U(x, \omega) \;\in\; \{0, \dots, m-1\}^{T}
\]
is the sequence of the first $T$ bytes the program emits, zero-padded if it
halts early.

As an example, the following program implements a three-iteration
\texttt{for} loop, using its first cell as the loop counter and emitting its
second cell once per iteration:
\[
  \underbrace{\texttt{+++}}_{\text{cell}_0 \,\gets\, 3}
  \quad
  \underbrace{\texttt{[}}_{\text{while cell}_0 \neq 0}
  \quad
  \underbrace{\texttt{>}\texttt{+}\texttt{.}}_{\text{cell}_1 \gets \text{cell}_1 + 1;\ \text{emit cell}_1}
  \quad
  \underbrace{\texttt{<}\texttt{-}}_{\text{cell}_0 \gets \text{cell}_0 - 1}
  \quad
  \underbrace{\texttt{]}}_{\text{end loop}}
  \quad
  \underbrace{\texttt{F}}_{\text{halt}}
\]
It emits $y = (1, 2, 3, 0, \dots, 0)$: one byte per iteration, then
zero-padding once the program halts.

Importantly, \emph{every} string over $\mathcal{A}$ is executable. There is
no syntax error: the only way a program can be malformed is through unmatched
brackets, which we treat as no-ops. 
In practice we cannot implement an unbounded tape or unbounded run times, so our
machine has finite memory and finite time budgets; wherever a program exceeds
one of these bounds, the semantics are defined to wrap or halt rather than
fault. 
The tape is circular, so that a head move off either end of the finite memory wraps around, and incrementing or decrementing a cell wraps modulo \(m\).
As a consequence, there are no out of bounds errors. 
Execution always terminates --- at the step budget, the end of the program,
or the $T$-th emitted byte, whichever comes first.

The machine as described is Turing complete using the eight
Brainf*ck instructions alone. 
In practice, however, common patterns such as clearing a
cell, moving a value to a neighbor, or scanning to the next zero cell are frequently used in human written Brainf*ck programs, and appear to increase the efficiency of self-play in preliminary experiments. We therefore extend the alphabet with the ten single-character tokens
in \cref{tab: macros}. 
All methods we compare draw from this same augmented alphabet, including the
Solomonoff-prior baseline in \cref{subsec: main-result} and the uniform-sampling
comparison, so the added primitives cannot account for any difference
between self-play and its controls. Several of the discovered program families in
\cref{tab:discovered-math-structure} use these tokens.

\begin{table}[ht]
\centering\small
\begin{tabular}{@{}cll@{}}
\toprule
Token & Expansion & Effect on tape \\
\midrule
Z & \texttt{[-]} & Clear current cell \\
R & \texttt{[->+<]} & Clear and add $x$ into right neighbor \\
L & \texttt{[->+++<]} & Clear and add $3x$ into right neighbor \\
N & \texttt{[-<->]} & Clear and subtract $x$ from left neighbor\\
C & \texttt{[->+>+<<]} & Clear and add $x$ into the two right cells \\
G & \texttt{[>]} & Scan right to next zero cell \\
H & \texttt{[<]} & Scan left to next zero cell \\
W & \texttt{[[-]>+<]} & If current $\neq$ 0: increment right and clear current \\
V & \texttt{[.>]} & Print stored string until a zero cell \\
X & \texttt{[-]++++++++++++++++} & Set cell to 16 \\
\bottomrule
\end{tabular}
\vspace{6pt}
\caption{The set of characters used to augment the Brainf*ck language, their expansion in terms of pure Brainf*ck, and their meaning. $x$ corresponds to the value at the current memory cell.}
\label{tab: macros}
\end{table}

\section{Reward Ablations Table}
To support our choice of reward we show several ablations along with variants of the real reward. The "none" reward is the canonical setup, which as we have seen is already better than the "uniform" ablation where we set the generator to sample program tokens uniformly from the alphabet. Additionally we consider a signed version of the reward which does worse than the absolute value, which shows that the absolute value is useful. When we shuffle the reward between programs of the same batch we break the correlation, and see a much worse result demonstrating that the mere marginal distribution of rewards is not sufficient to drive progress.

The last step reward is evaluating against the one-step weight difference rather than taking a window back to $e/2$ rounds, which is clearly worse, demonstrating that there is some value to averaging over larger blocks. Additionally the loss delta, which is putatively similar to the last step reward (to first order) is worse still, showing that the first-order reward is usefully more informative than the finite difference.

Finally the negative of the reward shows performance worse than a randomly initialized model, indicating that the reward does prefer systematically better programs to worse ones across the board.

\begin{table}[t]
\centering
\small
\setlength{\tabcolsep}{4.5pt}
\begin{tabular}{lccccccc}
\toprule
 & \multicolumn{7}{c}{ablation} \\
\cmidrule(lr){2-8}
dataset & None & uniform & signed & shuffle & last\_step$^{\,b}$ & loss\_delta$^{\,a,b}$ & negate \\
\midrule
text (dclm)                 & 5.34 & 7.75 & \textbf{5.06} & 5.96 & 6.39 & 7.40 & 10.62 \\
Metamath                    & \textbf{3.38} & 7.39 & 3.47 & 4.39 & 4.34 & 6.46 & 10.52 \\
C source                    & 4.16 & 7.92 & \textbf{4.10} & 4.79 & 4.95 & 6.37 & 10.60 \\
DNA (8-symbol)              & \textbf{2.29} & 3.02 & 2.45 & 2.50 & 3.22 & 3.57 & 8.37 \\
arithmetic                  & \textbf{0.22} & 7.94 & 0.51 & 0.73 & 1.26 & 1.92 & 10.64 \\
audio 8-bit PCM             & \textbf{2.27} & 3.93 & 2.45 & 2.91 & 3.39 & 3.67 & 7.67 \\
audio 16-bit PCM            & \textbf{5.02} & 6.08 & 5.24 & 5.79 & 5.97 & 6.32 & 9.02 \\
melody (Mutopia)            & \textbf{2.20} & 7.09 & 2.21 & 3.26 & 3.35 & 4.14 & 11.00 \\
CIFAR-10 (planar)           & \textbf{5.96} & 7.83 & 6.14 & 7.14 & 7.22 & 7.59 & 10.59 \\
random bytes         & \textbf{8.02} & \textbf{8.02} & 8.05 & \textbf{8.02} & 8.29 & 8.61 & 10.57 \\
\bottomrule
\end{tabular}
\vspace{2em}
\caption{
\textbf{Reward ablations of the self-play generator} at the 1M parameter model. Each entry shows the validation loss in bits per byte of the 4-seed ensemble on 256 held-out sequences per dataset, scored from the final-round checkpoints. Every ablation changes exactly one property of the canonical reward $r_i = {|\langle P\odot\nabla L(y_i;\theta_{\text{now}}),\, \theta_{\text{past}}-\theta_{\text{now}}\rangle|}$ with $\theta_{\text{past}}=\theta_{\lfloor e/2\rfloor}$ (\emph{None}). \emph{Signed} drops the absolute value, \emph{shuffle} permutes the rewards across the program pool, \emph{last\_step} uses the one-step window $\theta_{\text{past}}=\theta_{e-1}$, \emph{loss\_delta} uses the realized progress $L_i(\theta_{\text{pre}})-L_i(\theta_{\text{post}})$, \emph{negate} flips the reward's sign, and \emph{uniform} removes the generator entirely (i.i.d.\ uniform programs). On the majority of datasets, the canonical reward is best, or nearly best, and it often has lower variance as well. \\
$^{a}$ One loss\_delta seed stopped at round 2587 and is excluded; its ensemble is over 3 seeds. \\
$^{b}$ last\_step and loss\_delta are bimodal across seeds (e.g.\ dclm single-seed 6.4 / 6.5 vs 10.6 / 11.1 for last\_step; 7.0 vs 9.8 / 12.3 for loss\_delta); their ensembles average over the diverged seeds.
}
\label{tab:reward_ablations}
\end{table}

\section{Pool construction}
\label{sec:pool_construction}
At round $e$, we construct a fixed pool of programs
\[
  \mathcal{B}_e
  =
  \mathcal{B}_e^{\mathrm{fresh}}
  \mathbin{\dot\cup}
  \mathcal{B}_e^{\mathrm{mut}}
  \mathbin{\dot\cup}
  \mathcal{B}_e^{\mathrm{replay}}.
\]
The fresh pool $\mathcal{B}_e^{\mathrm{fresh}}$ consists of the latest programs sampled
from the generator $g_\phi$.
A fraction of on-policy
samples is replaced by $\mathcal{B}_e^{\mathrm{mut}}$: single-token
substitutions, insertions, or deletions of positively rewarded programs drawn
from the quality-diversity bank.
The replay pool $\mathcal{B}_e^{\mathrm{replay}}$ is formed by
drawing programs uniformly without replacement from the bank of
non-replay programs produced in earlier rounds. These programs are re-executed
with fresh random tapes. 
Mutations improve local exploration by making small edits to promising programs and replay mitigates catastrophic forgetting of behaviors discovered in earlier rounds.
For mutation, we use a MAP-Elites-style algorithm~\citep{mouret2015illuminatingsearchspacesmapping}, so that mutation
does not concentrate exclusively on the highest-reward programs. 
Concretely, each program is assigned to a niche according to two descriptors: (i) its maximum dynamic loop
depth, binned from $0$ through $8$ with larger depths clamped to the final bin,
and (ii) its program-body length, bucketed at $8$, $16$, and $32$ tokens.
This produces at most $36$ niches in total. 
Only programs receiving positive
generator reward are admitted to the archive, and within each niche we retain
the top $8$ programs according to their stored reward. Because the usefulness
of a program depends on the learner's current state, stored rewards are
decayed by a factor of $0.97$ each round, allowing newly useful programs to
replace stale elites. Mutation parents are selected uniformly across occupied
niches, rather than uniformly across all archived programs, preserving
representation for rarer structural behaviors such as deeply nested programs.

\section{Random-PCFG pretraining}
\label{app:pcfg}

\paragraph{Grammar sampling.}
Each grammar $G=(V,\Sigma,R,S)$ is drawn as follows. The terminal set
$\Sigma$ is a uniform sample, without replacement, of
$n_\Sigma\sim\mathcal{U}\{2,\dots,16\}$ distinct byte values from
$\{1,\dots,255\}$ (byte $0$ is reserved as padding and never emitted). The
grammar has $|V|\sim\mathcal{U}\{1,\dots,8\}$ non-terminals with start symbol
$S=V_0$. Each non-terminal receives $\mathcal{U}\{1,\dots,4\}$ productions;
production probabilities are i.i.d.\ $\mathcal{U}(0,1)$ weights (plus
$10^{-6}$), normalized to sum to one. Each right-hand side contains
$\mathcal{U}\{1,\dots,4\}$ symbols, each independently a terminal with
probability $p_T=0.5$ (uniform over $\Sigma$) and otherwise a uniform
non-terminal. \emph{Productivity repair:} if a non-terminal ends up with no
terminal-only production, the right-hand side of one uniformly chosen
production is replaced by a freshly sampled terminal-only string (its
probability unchanged), so every non-terminal can terminate in one step.

\paragraph{Derivation and row packing.}
A word is derived by leftmost expansion with an explicit stack, sampling
productions by their probabilities, and stops when the stack empties, the
output reaches $64$ bytes, or $10^4$ expansions elapse (a backstop for
grammars with unbounded expected yield); if expansion produces no terminal,
the start symbol's stored one-step terminal yield is emitted, so every word
has $\ge 1$ byte. For each
4{,}095-byte training row, \emph{one} fresh grammar is sampled and words are
derived from it and concatenated until the row fills (the final word is
truncated). Rows therefore contain repeated material from a single small
random grammar and are zero-free by construction.

\end{document}